\documentclass[journal,web]{article}
\usepackage{PRIMEarxiv}
\usepackage{graphicx}
\usepackage{eucal}
\usepackage{booktabs}
\usepackage{todonotes}
\usepackage{verbatim}
\usepackage{framed,multirow}
\usepackage{amsmath}
\usepackage{amssymb}
\usepackage{xcolor}
\usepackage{color, colortbl}
\usepackage[normalem]{ulem}
\useunder{\uline}{\ul}{}
\usepackage{changepage}
\usepackage{hyperref}
\definecolor{newcolor}{rgb}{.8,.349,.1}
\definecolor{Gray}{gray}{0.9}

\usepackage{todonotes}
\newcommand{\blue}[1]{\textcolor{black}{#1}}

\begin{document}

\title{Source-Free Domain Adaptation for~Image~Segmentation}

\author{
  Mathilde Bateson, Hoel Kervadec, José Dolz, Hervé Lombaert, Ismail Ben Ayed \\
  ETS Montréal \\
  \texttt{mathildebateson@gmail.com} \\
}


%

\maketitle

\newcommand{\mathbbm}[1]{\text{\usefont{U}{bbm}{m}{n}#1}}

\begin{abstract}
Domain adaptation (DA) has drawn high interest for its capacity to adapt a model trained on labeled source data to perform well on unlabeled or weakly labeled target data from a different domain. Most common DA techniques require concurrent access to the input images of both the source and target domains. However, in practice, privacy concerns often impede the availability of source images in the adaptation phase. This is a very frequent DA scenario in medical imaging, where, for instance, the source and target images could come from different clinical sites. We introduce a source-free domain adaptation for image segmentation. Our formulation is based on minimizing a label-free entropy loss defined over target-domain data, which we further guide with a domain-invariant prior on the segmentation regions. Many priors can be derived from anatomical information. Here, a class-ratio prior is estimated from anatomical knowledge and integrated in the form of a Kullback–Leibler (KL) divergence in our overall loss function.
Furthermore, we motivate our overall loss with an interesting link to maximizing the mutual information between the target images and their label predictions.
We show the effectiveness of our prior-aware entropy minimization in a variety of domain-adaptation scenarios, with different modalities and applications, including spine, prostate and cardiac segmentation. 
Our method yields comparable results to several state-of-the-art adaptation techniques, despite having access to much less information, as the source images are entirely absent in our adaptation phase. Our straightforward adaptation strategy uses only one network, contrary to popular adversarial techniques, which are not applicable to a source-free DA setting. Our framework can be readily used in a breadth of segmentation problems, and our code is publicly available: \url{https://github.com/mathilde-b/SFDA} 

\end{abstract}

\section{Introduction}
\subsection{Motivation}

Unprecedented advances in visual recognition tasks have been possible thanks to the improvements in hardware, novel deep architectures and availability of large annotated datasets. Deep Convolutional Neural Networks (CNNs) can provide powerful image representations when trained on huge amounts of labeled images, which can be used in a breadth of computer vision problems. For instance, CNNs have outstandingly improved automated methods for segmentation in many natural and medical imaging problems \cite{litjens2017survey}. A major impediment of such supervised models is that they require large amounts of training data built with scarce expert knowledge and labor-intensive, pixel-level annotations. Typically, segmentation ground truth is available for limited data, and supervised models are seriously challenged with new samples (target data) that differ from the labeled training samples (source data). In medical imaging, for instance, the data distribution may vary significantly across different vendors, machines, image modalities and acquisition protocols, as illustrated on Fig.~\ref{fig:s_t_im}. Such domain shifts between different scans introduce a significant variability in the appearances of the target regions, impeding the generalization of CNN segmentation models. 
There has been an ongoing research effort towards improving the performance of models across domains, without retraining them nor labeling entire datasets in new target domains, which would be impractical in medical imaging \cite{Cheplygina2018Notsosupervised}. 

Domain Adaptation (DA) addresses the transferability of a model trained on an annotated source domain to another target domain with no, or minimal annotations. With the advent of Generative 
Adversarial Networks (GANs) \cite{Goodfellow}, adversarial-learning techniques widely dominate the recent literature in domain adaptation for segmentation. One major limitation of adversarial 
techniques is that, by design, they require concurrent access to both the source and target data during the adaptation phase. More generally, other recent approaches to DA, such as those based on
self-training, also use both source and target data during adaptation. However, in many medical imaging scenarios, the source data may not be available in the adaptation phase. This involves, for example, 
confidentiality reasons, loss or corruption of the source data, or computational constraints for real-time applications.

\blue{Instead, we tackle \emph{Source-Free Domain Adaptation}, where the source data is not accessible during the adaptation phase. }
Our adaptation relies on minimizing a loss containing the Shannon entropy of predictions and a class-ratio prior on the target domain (i.e., the proportion of a region in an entire image). \blue{This loss implicitly matches the prediction statistics of the source and target domains, thereby removing the need for complex two-step adversarial training as in GANs. 
Moreover, we show the robustness of our framework to substantial uncertainty in the class-ratio prior, and give an information-theoretic perspective of our loss.
\textcolor{black}{Our method enables to embed approximate anatomical knowledge that is invariant across domains, and to leverage weak labels of the target samples in the form of image-level tags for segmentation tasks.}}
%


\subsection{Related Work}
\label{ssec:relatedwork}

Among the earliest works aiming to address domain-shift problems, \cite{Crammer2006Learning,ben2010theory,Pan2010Survey} propose to find a mapping of data distributions from a source to a target. 
More precisely, to tackle the discrepancy between the two domains, the learning process exploits the differences of data distributions across domains, yielding domain-invariant features. The main idea 
is to find an intermediate feature space where the marginal distribution of the source is similar to the target. Thus, we can assume that, in this intermediate representation, the prediction function 
is the same across source and target domains. This results in models that can be trained using annotated data sets from the source domain along with unlabeled or weakly labeled target data, with a strong 
cross-domain generalization ability.

\textbf{\textcolor{black}{Adversarial methods: }}Inspired by this assumption, recent works have focused on leveraging deep learning models to extract domain invariant features from input images \cite{ganin2015unsupervised,long2015learning,tzeng2015simultaneous}. 
Particularly, most of the existing research exploits deep adversarial training \cite{dann} in a wide range of applications and problems, such as classification \cite{ADDA,Wachinger2016Domain,vanTulder2016Representation,sankaranarayanan2017generate} or segmentation \cite{kamnitsas2017unsupervised,hoffman2017cycada,huo2018adversarial,javanmardi2018domain,tsai2018learning,zhang2018task,zhao2019supervised}. 
These methods either follow a generative approach, by transforming images from one domain to the other \cite{cyclegan,Huo2019}, or minimize the discrepancy in the feature or output spaces learnt by the model \cite{dou2018pnp,ADDA,tsai2018learning}. As these two perspectives are in essence complementary, the recent methods achieve state-of-the-art performances for adapting semantic segmentation in natural
\cite{hoffman2017cycada,Zhang2018FullyCA} and medical images \cite{SIFA} by combining image- and feature-alignment strategies. 
One major limitation of adversarial techniques is that, by design, they require concurrent access to both the source and target data during the adaptation phase. 

\begin{figure}[t]
    \includegraphics[width=1\linewidth]{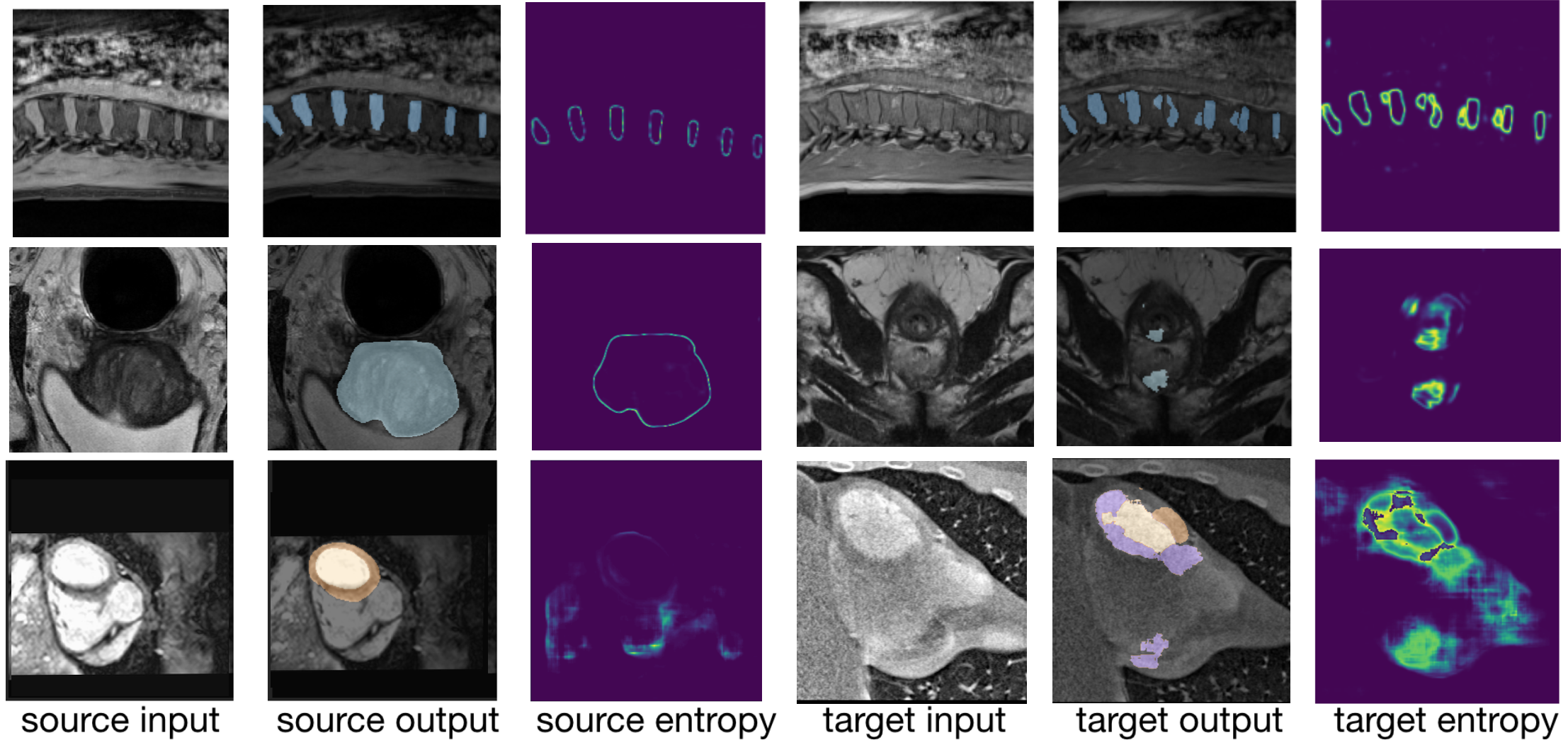}
      \caption[]{Visualization of severe domain shifts between source and target modalities along with their corresponding predicted segmentation and entropy maps in three applications. Top: 2 spine images from Water (left) and In-Phase (right) MRI, with the intervertebral disks depicted in blue and the background in black. Middle: 2 prostate MRI images from different sites. Bottom: 2 cardiac images from MRI (left) and CT (right). The cardiac structures of AA, LV and and MYO are depicted in blue, purple and brown, respectively. The domain shift in the target causes a drop in confidence and accuracy.}
         \label{fig:s_t_im}
\end{figure}

\textbf{\textcolor{black}{Self-training: }}Amongst alternative approaches to adversarial techniques, self-training \cite{zou2018unsupervised} and the closely-related entropy minimization \cite{advent,Wu2020EntropyMV,morerio2018minimalentropy} were investigated in computer vision. As confirmed by the low entropy prediction maps in Fig.~\ref{fig:s_t_im}, a model trained on an imaging modality tends to produce very confident predictions on within-sample examples, whereas uncertainty remains high on unseen modalities. Moreover, the entropy maps can identify inaccurate segmentation regions in these target examples.
\textcolor{black}{As a result, enforcing a higher confidence of predictions in the target domain would help decreasing this performance gap.} This is the underlying motivation for entropy minimization, which was first introduced in the contexts 
of semi-supervised \cite{Grandvalet} and unsupervised \cite{GomesNIPS2010} learning. 
To prevent the well-known collapse of entropy minimization to a trivial solution with a single class, the recent domain-adaptation methods in \cite{advent,Wu2020EntropyMV} further incorporate a criterion encouraging diversity in the prediction distributions, while \cite{uada} minimize the uncertainty measured as the variance of the network’s output, in combination with adversarial learning. 
However, similarly to adversarial approaches, all these uncertainty-based methods require access to the source data, both the images and labels, during the adaptation phase. The source data is used to compute the standard supervised cross-entropy loss and/or used in an adversarial adaptation, to prevent trivial solutions that are obtained by minimizing uncertainty on the unlabeled target images.  


\textbf{\textcolor{black}{Test-time Adaptation: }}\textcolor{black}{Closest to our work, test-time domain adaptation (TTA) was introduced to improve generalization to new and different data, possibly a single data point, at test times. Most TTA methods comply with the SFDA setting: they relieve the need for accessing source domain data after the source training phase. Initial SFDA attempts addressed adapting classification tasks \cite{sourcefree,sourcefree2}, either by using generative image translation \cite{Benaim2018} or self-supervision \cite{Sun2020,wang2021tent}. Extensions to segmentation problems \cite{KARANI2021101907,He2020,He2021} alter the source-domain training with auxiliary branches used to align the target and source domains in the pixel, network-feature, and/or network-output spaces. A drawback of these methods is that the source training phase is non-standard, and involves complex training schemes. \cite{Varsavsky} proposed a test-time adaptation based on domain adversarial learning, which is adapted to a single target-domain subject, but is not source-free.}

\textbf{\textcolor{black}{Domain Randomization}} \textcolor{black}{Recent work \cite{billot_learning_2020,synthseg} has investigated the possibility to segment scans of arbitrary contrasts and resolutions by training with synthetic intensity images. These methods, which only have been implemented on brain MRIs, also comply with the source-free domain adaptation scenario.} 

\textbf{\textcolor{black}{Weakly supervised segmentation in medical imaging:}}
\textcolor{black}{To alleviate the burden of pixel-wise annotation, weakly supervised learning has become a popular strategy. In this setting, the supervision received by the segmentation network may come in the form of image-level tags \cite{wu2019weakly,ouyang2019weakly,PATEL2022102374}, bounding boxes \cite{rajchl2016deepcut,kervadec2020bounding}, points \cite{khan2019extreme,dorent2021inter}, scribbles \cite{Tang2018Regularized}, target size \cite{jia2017constrained,kervadec2019constrained} or, more recently, shape descriptors \cite{KervadecMIDL2021}. 
On the one hand, approaches that rely on image-level tags typically use class-activation maps \cite{selvaraju2017grad}, which are deployed to generate pseudo-labels, mimicking fully-supervised learning. On the other hand, knowledge-driven approaches typically embed prior-knowledge, such as the target size or location, in the learning objective. Furthermore, while most prior literature relies on in-distribution data, a very few attempts investigated domain adaptation in a weakly-supervised setting \cite{Cheplygina2018Notsosupervised,Bateson2021,Paul_WeakSegDA_ECCV20,dorent2020scribble}. These works have shown promising results, especially when dealing with scarce data or severe domain shifts.}

\blue{\textbf{Leveraging the target class-ratio as a prior} has shown a great potential to guide the training of segmentation models when dealing with limited supervision, including weakly \cite{jia2017constrained,kervadec2019constrained}, semi-supervised \cite{yuille,KervadecMiccai} or few-shot \cite{boudiaf2021few} learning. In the presence of domain shifts, several recent works have also resorted to this prior as a source of additional supervision \cite{advent,zhang2019curriculum,Bateson2021}. An important difference, however, is that prior works require accessing the source data. Indeed, their learning objectives include a cross-entropy loss over the labeled source images during the training of the adaptation phase. This contrasts with our setting, as we relax this requirement.}






\subsection{Contributions}

We propose a \emph{Source-Free Domain Adaptation} formulation (SFDA) tailored to a setting where the source data is unavailable, neither its images nor its labeled masks, during the training of the adaptation phase. 
Instead, our method only requires the parameters of a model previously trained on the source data as an initialization; \textcolor{black}{ moreover, it does not use auxiliary branches or additional tasks trained on the source domain, contrary to previous SFDA methods \cite{KARANI2021101907,He2021,He2020}.}
Our formulation is based on a minimization of a label-free entropy loss defined over the target-domain data, which we further guide with a domain-invariant prior on the segmentation regions. \blue{To facilitate adaptation, we leverage weak supervision in the form of image-level tags in the target domain.}
Furthermore, we provide an interesting connection between our loss 
and the mutual information between the target images and their label predictions.

We report a comprehensive set of experiments and comparisons with state-of-the-art domain-adaptation methods, which shows the effectiveness of our prior-aware entropy minimization in three applications: the adaptation of spine segmentation across different MRI modalities, the adaptation of prostate segmentation in MRI modalities across different sites and machines, and the adaptation of cardiac segmentation from MRI to CT. Surprisingly, even though our method does not have access to the source data during adaptation, it achieves comparable or even better performances than several state-of-the-art methods \cite{zhang2019curriculum,tsai2018learning,ADDA,dann,cyclegan,dou2018pnp}, while greatly improving the confidence of network predictions. 

A preliminary conference version of this work has appeared at MICCAI~2020 \cite{Bateson2020}. \blue{This journal version provides (1) a new loss to tackle source-free adaptation, with an interesting mutual-information perspective and better gradient dynamics than the one introduced in \cite{Bateson2020}; (2) two new applications; (3) ablation studies; and (4) the introduction of anatomical knowledge to estimate the class-ratio priors, which demonstrates the practical usefulness of our method and its robustness to uncertainty in estimating the priors.} Specifically, unlike \cite{Bateson2020}, we perform comprehensive evaluations in a setting where the class-ratio priors of the target regions are not estimated by an auxiliary network, but rather derived from textbook anatomical knowledge, even with substantial imprecision. We argue that such an approach offers a great potential in multiple clinical settings, particularly when access to source data is compromised. Our framework can be readily used for adapting a breadth of segmentation problems, with the code made publicly available\footnote{\url{https://github.com/mathilde-b/SFDA}}. 
 
The contributions of this paper can be summarized as follows:
\begin{enumerate}
\item \textcolor{black}{We tackle Source-Free Domain Adaptation (SFDA), a setting where the source data is unavailable, neither its images nor labeled masks, during the training of the adaptation phase. Our formulation allows SFDA with no modification to the source training.}

\item We propose a novel loss defined over the unlabeled target-domain data, which integrates the Shannon entropy with a Kullback–Leibler divergence matching the class-ratios of the segmentation regions to an anatomical prior.
Furthermore, we motivate our loss with an interesting link to maximizing the mutual information between the target images and their latent labels.

\item We extensively validate our method on three DA datasets. The results show that our framework can effectively and efficiently address the domain shift problem without accessing the source data during the adaptation phase.
\end{enumerate}






\begin{figure*}[t]
    \includegraphics[width=1\linewidth]{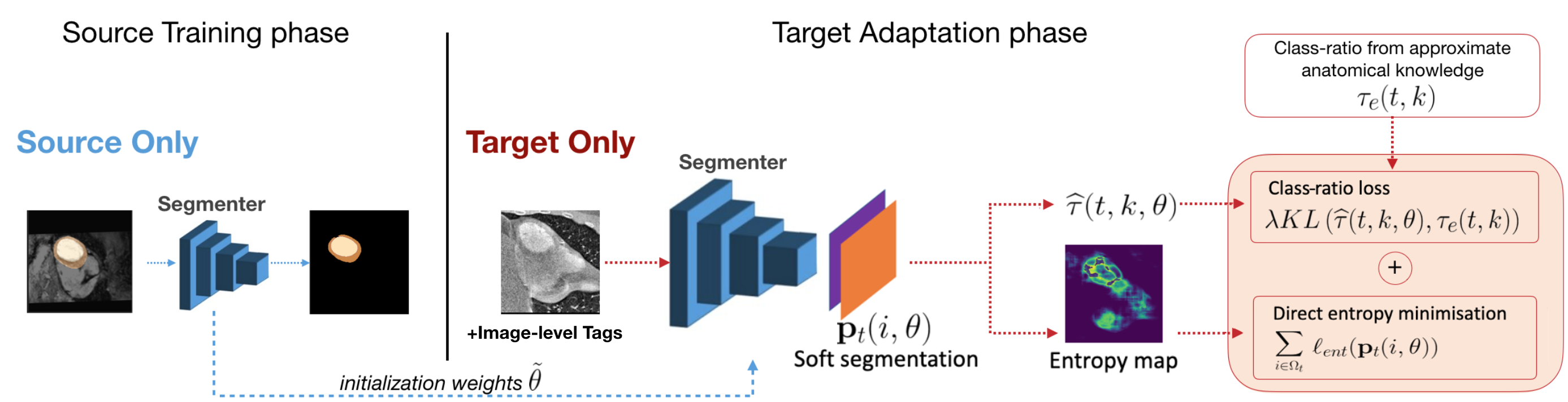}
      \caption[]{Overview of our framework for Source-Free Domain Adaptation: we leverage entropy minimization and a class-ratio prior, to remove the need for a concurrent access to the source and target data.}
      \label{fig:overview}
\end{figure*}

\section{Method} 
\label{sec:Method}
We consider a set $\mathcal{S}$ of source images ${I}_s: \Omega_s\subset \mathbb R^{d} \rightarrow {\mathbb R}$, $d\in \left \{ 2,3 \right \}$, $s=1, \dots, S$. The ground-truth $K$-class segmentation of $I_s$ can be written, for each pixel (or voxel) $i \in \Omega_s$, as a simplex vector ${\mathbf y}_s (i) = (y^1_s (i), \dots, y^K_s (i)) \in \{0,1\}^K$.
For domain adaptation (DA) problems, typically, a deep network is first trained on the source domain only, by minimizing a standard supervised loss with respect to network parameters $\theta$:
\begin{equation}\label{eq:source-sup}
\begin{aligned}
\mathcal{L}_{s}\left(\theta, \Omega_{s}\right)= \frac{1}{S}\sum_{s=1}^{S} \frac{1}{\left|\Omega_{s}\right|}\sum_{i\in \Omega_S}\ell\left({\mathbf y}_s (i), {\mathbf p}_s (i, \theta)\right)
  \end{aligned}
\end{equation}
where ${\mathbf p}_s (i, \theta) = (p^1_s (i,\theta), \dots, p^K_s (i, \theta)) \in [0,1]^K$ is the softmax output of the network at $i$ in image $I_s$, \textcolor{black}{and here we take $\ell$ as the standard cross-entropy loss} : 
$\ell({\mathbf y}_s (i), {\mathbf p}_s (i, \theta)) = - \sum_k y^k_s (i) \log p^k_s (i, \theta)$.

The adaptation phase is then initialized with the network parameters $\tilde{\theta}$ obtained from the source training phase. Given a set $\mathcal{T}$ of images in the target domain, ${I}_t: \Omega_t\subset \mathbb R^{2,3} \rightarrow {\mathbb R}$, $t=1, \dots, T$,
the first loss term in our adaptation phase encourages high confidence in the softmax predictions 
of the target, which we denote ${\mathbf p}_t (i, \theta) = (p^1_t (i,\theta), \dots, p^K_t (i, \theta)) \in [0,1]^K$. This is done by minimizing a weighted Shannon entropy of each of these predictions: 
\begin{equation}
\label{eq:ent-target}
\ell_{ent}({\mathbf p}_t (i,\theta)) = - \nu_k\sum_k p^k_t (i,\theta) \log p^k_t (i, \theta)
\end{equation}
where $\nu_k,k=1, \dots, K$, are non-negative constants denoting class weights added to alleviate the burden of unbalanced class-ratios.

However, it is well-known from the semi-supervised and unsupervised learning literature \cite{Grandvalet,GomesNIPS2010,JabiPAMI} that minimizing this entropy loss alone may result into trivial solutions, where the predictions are biased towards a 
single dominant class. To avoid such degenerate solutions, the recent domain-adaptation work of \cite{advent,Wu2020EntropyMV} have integrated a standard supervised cross-entropy loss over the source data, such as in Eq.~\eqref{eq:source-sup}, when training during the adaptation phase. This, however, requires access to the source data, both its images and labels, during the adaptation phase. To remove this undesired requirement, we embed a domain-invariant prior knowledge to guide the unsupervised entropy training during the adaptation phase, which takes the form of a class-ratio prior (i.e., the proportion of a region in an entire image). The unknown true class-ratio prior for a class $k$ and image $I_t$ can be computed as follows: $\tau_{GT}(t,k) = \frac{1}{\left|\Omega_{t}\right|}\sum_{i \in \Omega_t}  y^k_t (i)$. This gives the size of class $k$ in image $I_t$ over the image size. However, as the ground-truth labels are unavailable in the target domain, this prior cannot be computed directly. 
Instead, we estimate it with simple region statistics from anatomical prior knowledge, which we denote as $\tau_{e}(t,k)$. 
Furthermore, the class-ratio of the segmentation network output prediction can be computed as follows: $\widehat{\tau}(t,k,\theta)=\frac{1}{\left|\Omega_{t}\right|} \sum_{i \in \Omega_t} p^k_t (i, \theta)$.

We regularize the entropy in Eq.~\eqref{eq:ent-target} with a Kullback-Leibler (KL) divergence matching these two class-ratios. Thus, our method minimizes the following overall
loss during the training of the adaptation phase:
\begin{equation}\label{eq:AdaMI}
\begin{aligned}
  \min_{\theta}\sum_{t} \frac{1}{\left|\Omega_{t}\right|} \sum_{i \in \Omega_t} \ell_{ent}({\mathbf p}_t (i, \theta))+ \mbox{KL}(\widehat{\tau}(t,\theta,\cdot),\tau_{e}(t,\cdot))
  \end{aligned}
\end{equation}

\textcolor{black}{where $\mbox{KL}(\widehat{\tau}(t,\theta,\cdot),\tau_{e}(t,\cdot))= \widehat{\tau}(t,\theta,\cdot)\log\left ( \frac{\widehat{\tau}(t,\theta,\cdot)}{\tau_{e}(t,\cdot)} \right )$.}

Clearly, minimizing our overall loss in Eq.~\eqref{eq:AdaMI} during adaptation does not use the source images and labels. In the following, we discuss an interesting link between our loss in Eq.~\eqref{eq:AdaMI} and maximizing the mutual information between the target images and their network predictions. Figure~\ref{fig:overview} shows the overview of the proposed framework.  

\subsection{\textcolor{black}{Link to mutual-information maximization}}

Notice that the terms of the KL penalty in Eq.~\eqref{eq:AdaMI} are inverted compared to our initial formulation (\emph{AdaEnt}), which we provided in the conference version of this work \cite{Bateson2020}; see Eq.~\eqref{eq:AdaEnt}. 
Besides the empirical motivation (as it will be shown in the experimental section hereafter), this is first and foremost motivated by theoretical results in information theory, as we link below Eq.~\eqref{eq:AdaMI} to maximizing the mutual information between the input images and their latent label predictions. The full proof is derived in \ref{miproof}.

Let $\mathcal{I}(X,Y)$ denote the mutual information between two random variables $X$ and $Y$:
\begin{equation}
\begin{aligned}
\mathcal{I}(X ; Y) &=H(Y)-H(Y \mid X) \\
&=-\mathbb{E}_{Y}\left[\log \mathbb{E}_{X}[p(Y \mid X)]\right]+\mathbb{E}_{X, Y}[\log p(Y \mid X)]
\end{aligned}
\end{equation}
where $H(Y)$ is the entropy of $Y$, $H(Y\mid X)$ is the conditional entropy of $Y$ given $X$, and $\mathbb{E}_{X}[p(Y \mid X)]$ is the marginal distribution of Y under the conditional model $p(Y \mid X)$. 

We denote ${P}_t$ the $K \times |\Omega_t|$ softmax prediction mask, i.e. matrix whose columns are the vectors of network outputs ${\mathbf p}_t (i, \theta), i \in \Omega_t $. Given the classical interpretation of the softmax predictions as probabilities: $p_t^k(i,\theta)=p(y_t^k(i)=1\mid I_t,\theta)$, the empirical class-ratio distribution is an estimate of the marginal distribution of ${P}_t$: $\hat{\tau}(t,\theta,\cdot) = \frac{1}{\left | \Omega_t \right |} \sum_{i \in \Omega_t} {\mathbf p}_t (i, \theta) = \mathbb{E}_{I_t}[p(P_t \mid I_t)]$.
Therefore the empirical estimate of the mutual information between the images $I_t$ and their softmax predictions, $P_t, t=1, \dots, T$, can be expressed as\footnote{See details of proof in \ref{miproof}}:

\begin{equation}
 \mathcal{I}_\theta=\frac{1}{T}\sum_{t}\underbrace{H\{\hat{\tau}(t,\theta,\cdot)\}}_{-\mathbb{E}_{P_t}\left[\log \mathbb{E}_{I_t}[p(P_t \mid I_t)]\right]}- \underbrace{\frac{1}{\left|\Omega_{t}\right|}\sum_{i \in \Omega_t} \ell_{ent}({\mathbf p}_t (i, \theta))}_{-\mathbb{E}_{I_t,P_t}[\log p(P_t \mid I_t)]}
\label{eq:MI}
\end{equation}

In the different context of discriminative clustering, \cite{GomesNIPS2010} draw a connection between maximizing the empirical estimate of the mutual information, as in Eq.~\eqref{eq:MI}, and a generalization of the mutual information based on the KL divergence, as in Eq.~\eqref{eq:AdaMI}.
Indeed, note that the following basic identity holds:
\begin{equation}
\begin{aligned}
H\{\hat{\tau}(t,\theta,\cdot)\}\stackrel{\mathrm{c}}{=} - KL\{\hat{\tau}(t,\theta,\cdot), U\}
  \end{aligned}
\end{equation}
where $U$ is the uniform distribution over labels $\{1, \cdots, K\}$. The term $KL\{\hat{\tau}(t,.,\theta), U\}$ is maximized when the class-ratio distribution is uniform. Instead, to integrate a prior about the class-ratio distribution, for each image $I_t$ and class $k$, we can replace $U$ by prior distribution $\tau_{e}(t,.)$ as follows:
\begin{equation}
\max_{\theta} \sum_{t}-KL\{\hat{\tau}(t,\theta,\cdot), \tau_{e}(t,\cdot)\}-\sum_{t}\frac{1}{\left|\Omega_{t}\right|}\sum_{i \in \Omega_t} \ell_{ent}({\mathbf p}_t (i, \theta)
\label{eq:F}
\end{equation}
which is equivalent to Eq. \eqref{eq:AdaMI}. Maximizing the mutual information between the images $I_t$ and their softmax predictions $p_t(\theta)$ is a principled approach in unsupervised problems, such as unsupervised discriminative clustering \cite{GomesNIPS2010, JabiPAMI}, further motivating our formulation, which we denote \textit{AdaMI} in the following.


\subsection{Choosing the penalty function}\label{ssec:invert}

\begin{figure}[t]
\centering
    \includegraphics[width=0.5\linewidth]{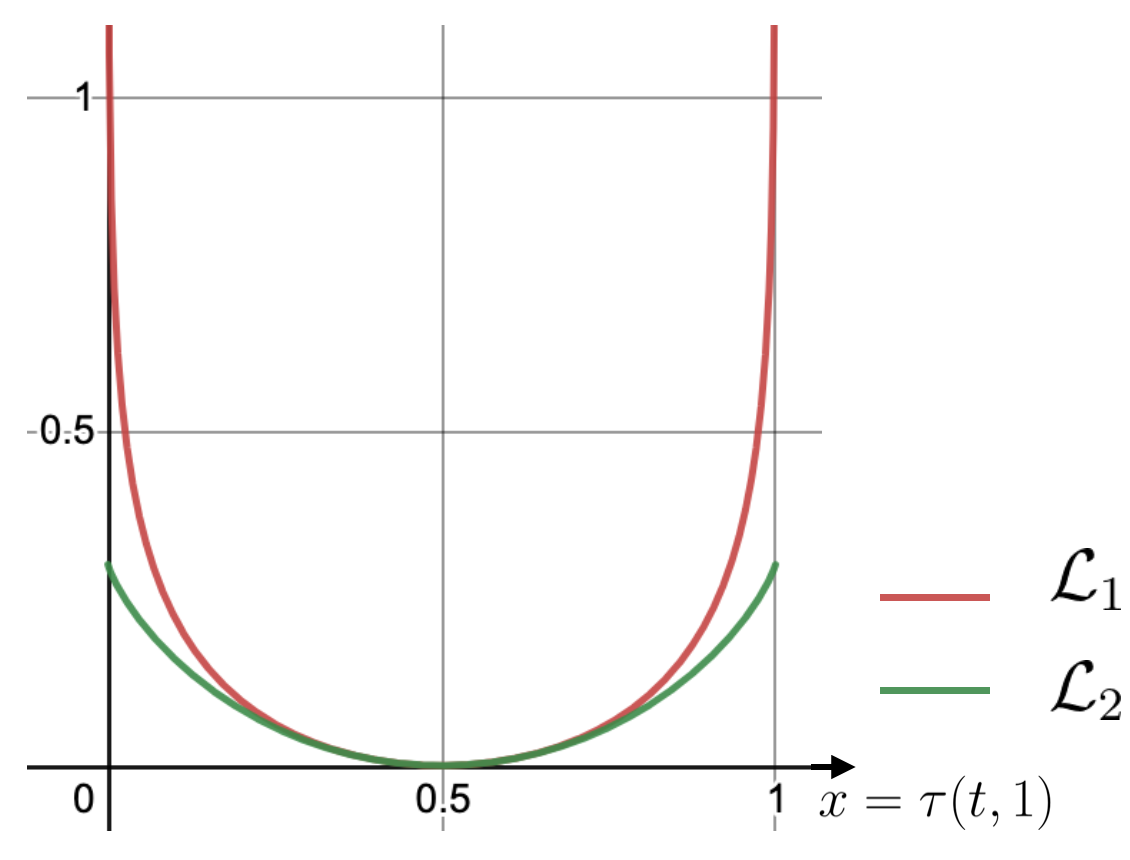}
      \caption[]{Comparison of two class-prior losses in the scenario $K=2$, with the ground-truth class-ratio set to $\tau_{GT}(t,1) = 0.5$: The plots illustrate better gradient dynamics of $\mathcal{L}_2$ at the vicinity of a class-ratio $\tau(t,1) = 0$. Best seen in colors.}
         \label{fig:loss_graph}
\end{figure}

Given an image $I_t$, consider the penalty functions $\mathcal{L}_1$ (resp. $\mathcal{L}_2$) used in combination with entropy minimization in \textit{AdaEnt} (resp. in \textit{AdaMI}) :
\begin{equation*}
\begin{split}
  \mathcal{L}_1 = \mbox{KL}(\tau_{e}(t,\cdot),  \widehat{\tau}(t,\theta,\cdot)) \\
\mathcal{L}_2 = \mbox{KL}(\widehat{\tau}(t,\theta,\cdot), \tau_{e}(t,\cdot)) 
\end{split}
\end{equation*}
Figure \ref{fig:loss_graph} shows the profile of these two regularizers as functions of the class-ratio for a binary-segmentation case, with a target foreground class-ratio set to 0.5. We see that $\mathcal{L}_2$ may be a better choice than $\mathcal{L}_1$ when the initial predictions of the network are extremely imbalanced. \textcolor{black}{Indeed, note the gradient properties and stability at the vicinity of 0, i.e., when the predicted foreground class-ratio $\tau(t,1)$ is close to 0. We see that both first and second derivatives of the regularizer are unbounded for $\mathcal{L}_1$, but bounded and constant for $\mathcal{L}_2$.} Our experiments confirm the superiority of the $\mathcal{L}_2$ regularizer, in terms of training stability and quantitative performance.

\subsubsection{Estimating the class-ratio prior from anatomical knowledge}\label{ssec:sizeprior}

In \cite{Bateson2020}, the ground-truth class-ratio is estimated through an auxiliary network trained with the source data. In a more general source-free scenario, only the weights $\tilde{\theta}$ of a network trained with the source data are available during the adaptation phase, and the class-ratio cannot be learnt, neither estimated from the source data. Therefore, we resort here to the more general case where the true class-ratio $\tau_{GT}(t,k)$ of each structure $k$ in an image $I_t$ is estimated from anatomical knowledge $\bar{\tau}_k$ available in the clinical literature (see \ref{appendix} for our estimates from anatomical information). 

For each 2D target image $I_t$ and each structure $k$, the class-ratio used for adapting the segmentation network with Eq.~\eqref{eq:AdaMI} is obtained by adding weak supervision in the form of image-level tag information:

 \begin{equation}
\tau_{e}(t,k)=\begin{cases}
\bar{\tau}_k & \text{if region $k$ is within image $t$}.\\
    0 & \text{otherwise},
  \end{cases}
  \label{eq:constraint_lit}
\end{equation}

Note that we use exactly the same class-ratio priors and weak supervision in our \textit{AdaEnt} method, for a fair comparison.

\section{Experiments and Results}\label{sec:exp}

\subsection{Experimental Settings} 

\subsubsection{Data sets}\label{ssec:dataset} 

\paragraph{\textbf{IVDM3Seg}}The proposed SFDA method is first evaluated on the dataset from the MICCAI 2018 IVDM3Seg Challenge\footnote{https://ivdm3seg.weebly.com/}, consisting of 16 3D multi-modality MRI data sets, collected from 8 subjects at two different stages to study inter-vertebral disc (IVD) degeneration. The scans were generated by a Dixon protocol with a $1.5 \mathrm{~T}$ Siemens MRI scanner, producing four aligned modalities. \textcolor{black}{Scans are acquired in sagittal direction.} Each volume has an anisotropic resolution of $2 \times 1.25 \times 1.25 \mathrm{~mm} / \mathrm{vx}$. The corresponding manual segmentations of the IVDs are also available.  In our experiments, we set the water modality (Wat) as the source and the in-phase (IP) modality as the target domain. 
Therefore, in this setting, the source and target modalities are acquired from the same patient.  \textcolor{black}{From this dataset, 12 scans are used for training, one for validation, and the remaining 3 scans for testing}.  \textcolor{black}{Images are normalized to zero mean and unit variance. Then, we performed a data augmentation based on affine transformations.} The setting is binary segmentation (K=2).

\paragraph{\textbf{NCI-ISBI13}}
We employ prostate T2-weighted MRIs from
2 different data sources with distribution shifts from the NCI-ISBI13 dataset, with their corresponding manual segmentations of the prostate region. The source dataset consists of 30 volumes from Radboud University Nijmegen Medical Centre, generated with a $3 \mathrm{~T}$ Siemens scanner. The target dataset consists of 30 volumes from Boston Medical Center generated with a $1.5 \mathrm{~T}$ Philips Achieva. We use the publicly available pre-processed data provided by \cite{SAML}, which resized each sample to $384\times384$ in axial plane, normalized it to zero mean and unit variance. \textcolor{black}{We employed data augmentation based on affine transformations.} \textcolor{black}{We use 19 scans for training, one for validation, and the remaining 10 scans for testing.} 

\paragraph{\textbf{MMWHS}}
We employ the 2017 Multi-Modality Whole Heart Segmentation (MMWHS) Challenge dataset for cardiac segmentation \cite{Zhuang2019}. The dataset consists of 20 MRI (source domain $S$) and 20 CT volumes (target domain $T$) of non-overlapping subjects, with their corresponding ground-truth masks. We adapt the segmentation network for parsing four cardiac structures: the Ascending Aorta (AA), the Left Atrium blood cavity (LA), the Left Ventricle blood cavity (LV) and the Myocardium of the left ventricle (MYO). We employ the pre-processed data provided by \cite{dou2018pnp}, as well as their data split, \textcolor{black}{with 14 subjects used for training, 2 for validation, and 4 for testing}. \textcolor{black}{All the data were normalized as zero mean and unit variance}. In order to obtain a similar field of view for all volumes, they cropped the original scans to center the structures to segment using a 3D bounding box with a fixed coronal plane size of $256\times256$. Then, they performed a data augmentation based on affine transformations. We use this augmented dataset for our proposed method as well as the benchmark methods that we implemented.



\subsubsection{Benchmark Methods}\label{ssec:benchmark}

Quantitative evaluations and comparisons with state-of-the-art methods are reported hereafter.
We compare our proposed model \textit{AdaMI} to the benchmark methods below, which have shown state-of-the-art performances for adapting segmentation networks.

\textit{Source-Free AdaEnt:} We compare to the loss that we proposed in our original source-free domain adaptation \cite{Bateson2020}, denoted $AdaEnt$ in the following:
\begin{equation}
\sum_{t} \frac{1}{\left|\Omega_{t}\right|}\sum_{i \in \Omega_t} \ell_{ent}({\mathbf p}_t (i, \theta))+ \lambda \mbox{KL}(\tau_{e}(t,\cdot),\widehat{\tau}(t,\theta,\cdot))
\label{eq:AdaEnt}
\end{equation}

\textit{Constrained Domain Adaptation:} We compare to the method adopted in \cite{Bateson2021}, referred to below as \textit{CDA}:
\[\mathcal{L}_{s}\left(\theta, \Omega_{s}\right)+ \frac{\lambda}{T} \sum_{t=1}^{T} [\tau_{e}(t,\cdot)-\widehat{\tau}_{t}(t,\theta,\cdot)]^2\]

\textit{Curriculum Domain Adaptation:} We denote \textit{AdaSource} the method adopted in \cite{zhang2019curriculum}:
\[\mathcal{L}_{s}\left(\theta, \Omega_{s}\right)+ \frac{\lambda}{T} \sum_{t=1}^{T} KL(\tau_{e}(t,\cdot),\widehat{\tau}_{t}(t,\theta,\cdot))\]

\textit{Adversarial Domain Adaptation:} We compare to \textit{AdaptSegNet}, the method adopted in \cite{tsai2018learning}:
\[\mathcal{L}_{s}\left(\theta, \Omega_{s}\right)- \frac{\lambda}{T} \sum_{t=1}^{T}\sum_{i \in \Omega_T} \log \left(D\left(p_{t}\right(i,\theta))^{(1)}\right)\]
where the adversarial loss maximizes the probability of a target sample being predicted as the source by a discriminator $D$.

Note that, for \textit{CDA}, \textit{AdaSource} and \textit{AdaptSegNet}, the images from the source and target domains must be present concurrently during the adaptation phase. 
\textcolor{black}{For \textit{CDA} and \textit{AdaSource}, the class-ratio is estimated through an auxiliary network trained with the source data and the weakly-supervised target data, as in \cite{Bateson2020}}. 

\textcolor{black}{We also compared to the following two source-free domain adaptation methods. The first is TTA \cite{KARANI2021101907}, which trains an auxiliary denoising autoencoder on the source, then applies it to the noisy segmentations in the target. The second is Tent \cite{wang2021tent}, which uses a simple entropy minimization, similarly to Eq.~\eqref{eq:ent-target}. Importantly, for both methods, instead of optimizing the whole segmentation network, only the normalization statistics and affine parameters of the network are updated, while the rest of the parameters are freezed.}

A model trained on the source domain only using Eq.~\eqref{eq:source-sup}, \textit{NoAdap}, is used as a lower bound. A model trained with the supervised cross-entropy loss on the target domain, referred to as $Oracle$, serves as an upper bound.

\textcolor{black}{Finally, for the cardiac application, we also present benchmark results obtained in previous DA works (\cite{uada,dou2018pnp}), which we directly report in Table \ref{table:resultswhs2}. The methods using AdaNet as the backbone were implemented in \cite{dou2018pnp}, those with DeepLabV2 were implemented in \cite{uada}.}


\subsubsection{Evaluating robustness to class-ratio prior imprecision}
\label{sssec:robustness}

In the following experiments, we investigate the impact on our SFDA approach of both precise and imprecise prior information about the class-ratios in the target domain. 
To this end, we train several models under the same setting, validating different values for the class-ratio priors on the target images. We illustrate on the challenging problem of segmenting cardiac structures, which have a high class-ratio variance amongst slices.

First, we investigate the capability of SFDA in the ideal setting when the precise size of the segmented region is known. To this end, for each image $t$ and each structure $k$ of the target domain, we use the following class-ratio derived from the ground-truth size:
\begin{equation}
    \tau_{GT}(t,k) = \frac{1}{\left|\Omega_{t}\right|}\sum_{i \in \Omega_t}  y^k_t (i)
\end{equation}

This setting is hereafter referred as $AdaMI_{\tau_{GT}}$. 
This is followed by evaluating the robustness of our benchmarked method to a varying imprecision of the prior knowledge on the class-ratio prior, i.e., varying the size estimates of the segmented regions.\blue{For each image $t$ and each structure $k$ of the target domain (except the background), we use the following error on the class-ratio}:  

\begin{equation}
    \tau(t,k) = (1\pm \delta)\tau_{e}(t,k)
\end{equation}

\blue{And then obtain the estimate the background estimation as : $\tau(t,0) = 1 -\sum_{k>1}\tau(t,k)$.} We validate using imprecision errors varying with $\delta$: $\left \{ 0.2,0.4,0.6 \right \}$ and denote this setting $AdaMI_{\delta\tau}$ below. 

\subsubsection{\textcolor{black}{Ablation studio on target training dataset size}}
\label{sssec:sizetargettrain}
\textcolor{black}{In this experiment, we study how much target training data is necessary for our method to achieve a successful adaptation. We train several models under the same setting, with a varying number of subjects in the target training dataset. This setting is hereafter referred as $AdaMI_{NT1},AdaMI_{NT2}..$}

\subsubsection{\textcolor{black}{Ablation studio on the weak annotations in the target training dataset}}
\label{sssec:notag}
\textcolor{black}{Finally, we investigate the impact of removing the image-level tags in the target training dataset, i.e. a fully unsupervised source-free DA setting. Instead, we use an \textit{estimation} of this tag derived from the network prediction, and select a \textit{subset} of the target training dataset. More specifically, for each 2D target training image $I_t$ and each structure $k$:}

\blue{ \begin{equation}
I_t\text{ is }\begin{cases}
\text{selected, with }\tau_e (t,k) =\bar{\tau}_k& \text{if $\widehat{\tau}(t,\tilde{\theta},k)>\frac{1}{4}\bar{\tau}_k$}.\\
\text{selected, with }\tau_e (t,k) =0& \text{if $\widehat{\tau}(t,\tilde{\theta},k)=0$}.\\
    \text{discarded otherwise }
  \end{cases}
  \label{eq:constraint_notag}
\end{equation}}

\blue{With $\tilde{\theta}$ the initial network parameters at the start of the adaptation phase. We then update this estimation once during training, at the epoch 100.}

\subsubsection{Training and implementation details}
\label{ssec:implementation}
For all the methods, we employed UNet \cite{UNet}, a widely used segmentation network due to its simplicity. \textcolor{black}{The architecture used is the same one as for the original UNet paper. We use a 2D implementation for all applications}. \blue{In the source training phase, a model is trained on the source data only with Eq.~\eqref{eq:source-sup} for 150 epochs, a learning rate of $5\times10^{-4}$, and a learning rate decay of 0.9 every 20 epochs. The final model is used as initialization to the adaptation phase, when the model is adapted with Eq.~\eqref{eq:AdaMI}, trained with the Adam optimizer \cite{Adam}, for 150 epochs. For all applications, the initial learning rate is $1\times10^{-6}$, the weight decay is $10^{-3}$, and the batch size is 24. The learning rate decay is 0.7 for the heart and prostate applications, and 0.2 for the spine one. It is applied every 20 epochs. For all methods, we pick the final model as the one at the last epoch.} The weights from Eq.~\eqref{eq:ent-target} are calculated as: $\nu_k=\frac{\bar{\tau_k}^{-1}}{\sum_k \bar{\tau_k}^{-1}}$.



\subsubsection{Evaluation metrics}
\label{ssec:evaluationmet}
\textcolor{black}{Our first evaluation metric is the Dice similarity coefficient (DSC), which measures the voxel-wise segmentation accuracy between the predicted and reference volumes. The second is the average symmetric surface distance (ASD), which calculates the average distances between the surface of the prediction mask and the ground truth.} 
As the data is volumetric for all applications, these metrics are computed over the 3D segmentation masks.

\subsection{Quantitative results}
\label{ssec:quant}
The quantitative performances of the different methods are presented in Table~\ref{table:resultsivd} for the spine and prostate images, and in Table~\ref{table:resultswhs2} for the cardiac images. 

\begin{table}[h!]
\centering
\footnotesize
\caption{Performance comparison of the proposed formulation with different domain adaptation methods for spine (IVDM3Seg dataset, left) and prostate (NCI-ISBI13 dataset, right) segmentation, in terms of DSC ($\%$) and ASD (vox).}
  \setlength\tabcolsep{1.pt}
\vspace{3mm}
\begin{tabular}{lcc|cc|cc}
\midrule
\textbf{Method} & \textbf{Source} &  \textbf{Target} & \multicolumn{2}{c}{\textbf{Spine IVDs}} &\multicolumn{2}{c}{\textbf{Prostate}}\\
 & \textbf{Free} & \textbf{Tags} &  \textbf{DSC} & \textbf{ASD} & \textbf{DSC} & \textbf{ASD}\\
\midrule
NoAdap (lower bound) &$\checkmark$  &  $\times$   &68.5 & 2.15 & 67.2 & 10.59\\
Oracle (upper bound)  &$\checkmark$ &  $\checkmark$   & 87.5  & 0.38  & 88.4 & 1.81\\
\midrule 
AdaptSegNet \cite{tsai2018learning} &$\times$& $\times$     & \textbf{82.4}  & \textbf{0.50}  & \textbf{83.1} & \textbf{2.43} \\
\begin{tabular}[c]{@{}l@{}}AdaSource \cite{zhang2019curriculum}\end{tabular}  &$\times$& $\checkmark$& 75.9 &  0.99  & 76.3 & 3.93\\
\begin{tabular}[c]{@{}l@{}}CDA \cite{Bateson2021}\end{tabular} &$\times$  &$\checkmark$ &  75.7 & 0.86 & 77.9 & 3.28\\
\begin{tabular}[c]{@{}l@{}}TTA \cite{KARANI2021101907}\end{tabular} &$\checkmark$  & $\times$ & 69.7  & 1.65 & 73.2 & 3.80\\
\begin{tabular}[c]{@{}l@{}}Tent \cite{wang2021tent}\end{tabular} &$\checkmark$  &$\times$ &  68.8 &  1.84 & 68.7 & 5.87\\
\rowcolor{Gray}\begin{tabular}[c]{@{}l@{}}Prior AdaEnt \cite{Bateson2020}\end{tabular}  &  $\checkmark$& $\checkmark$ & 72.9  & 1.54 & 77.8 & 4.10\\
\rowcolor{Gray}\begin{tabular}[c]{@{}l@{}}AdaMI (Ours)\end{tabular} &  $\checkmark$& $\checkmark$&  74.2 & 1.17 & 79.5 & 3.92\\
\midrule
\end{tabular}
\label{table:resultsivd}
\end{table}

\begin{figure}[t]
  \centering
  \includegraphics[width=0.6\linewidth]{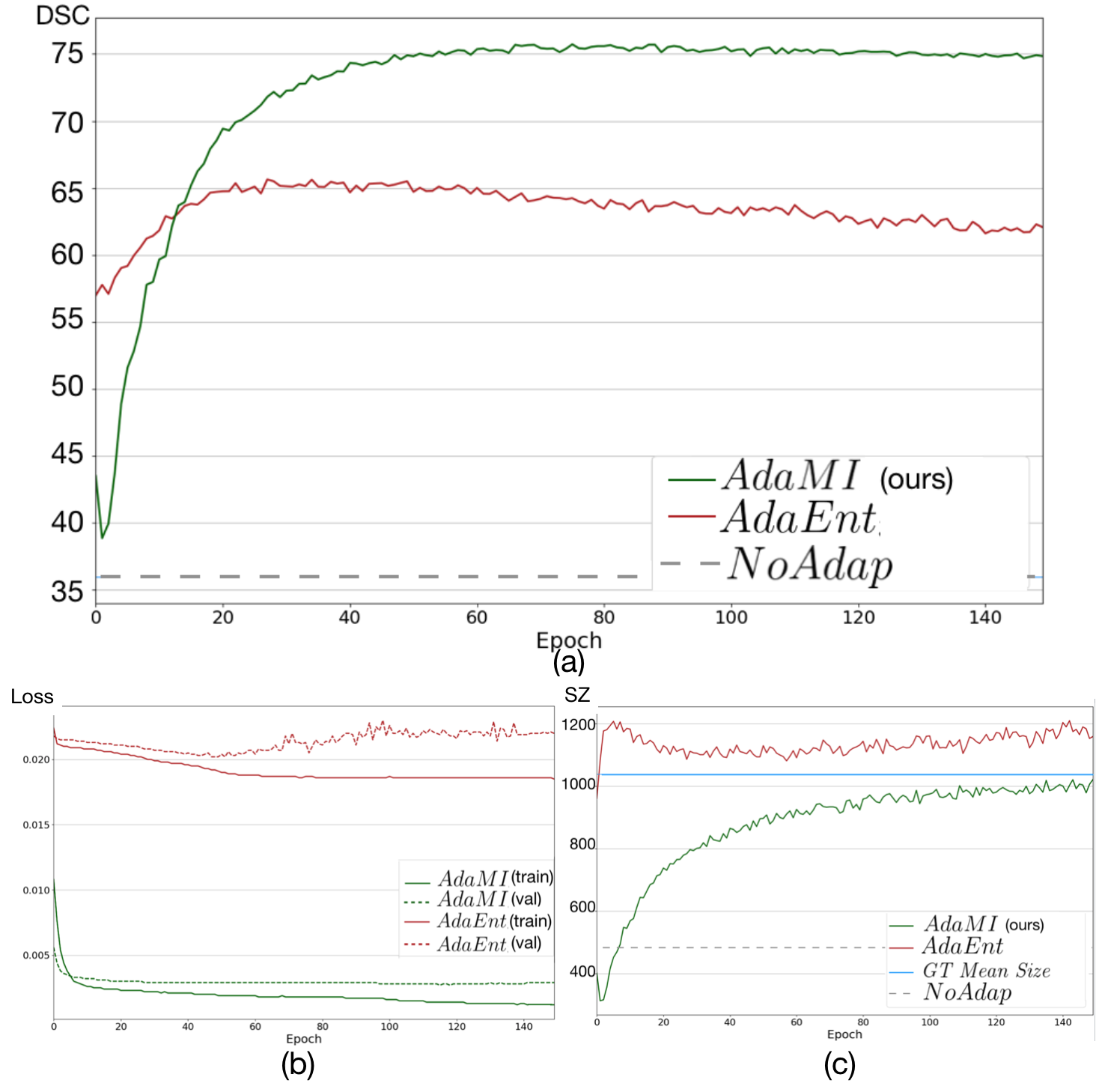}
\caption[]{Quantitative performance: (a) Evolution of DSC (\%) and (b)\blue{ Learning Curves and} (c) mean ground truth sizes and predicted sizes (px) of cardiac structures segmentation masks over training epochs on target images from the validation set. Comparison of the proposed model \textit{AdaMI}, and our previous \textit{AdaEnt}.}
\label{fig:diceszfig}
\end{figure}

\paragraph{\textcolor{black}{No Adaptation}}First, we see that the models trained with full supervision on the source domain suffer from a drop in performance when used in a different target domain without any adaptation. In Fig.~\ref{fig:diceszfig}(c), it can be verified that the \textit{NoAdap} is in an under-segmentation regime, with the predicted sizes of structures well below their true sizes. This validates that the predictions are biased towards the dominant class, which is the background here. 
\paragraph{\textcolor{black}{With Adaptation}}
All models that use adaptation yield a substantial improvement over the lower baseline. \blue{For instance, on spine images, our model \textit{AdaMI} reaches a Dice score (DSC) of 74.2\%, representing 90\% of the best-performing adaptation method, \textit{AdapSegNet} \cite{tsai2018learning}, which used the source data during adaptation. \textit{AdaMI} yields a 1.17 ASD, which corresponds to an improvement by a multiplicative factor of 1.8 compared to the value for \textit{NoAdap} (2.15 ASD). On prostate images, \textit{AdaMI} reaches 79.5\% DSC, 95\% of the top performance \textit{AdapSegNet}. An ASD of 3.92 is obtained, an improvement by a multiplicative factor of 3 compared to the value for \textit{NoAdap} (10.59 ASD). Surprisingly, on cardiac images, where the domain shift is higher, \textit{AdaMI} ranks second out of sixteen other adaptation techniques in terms of average DSC across cardiac structures, outperformed only by the recent method in \cite{uada}, a substantially more complex adaptation framework. Note that the quantitative results are not directly comparable between all models, since the backbone networks differ (see Table \ref{table:resultswhs2}).} 
These results show that having access to more information on source data does not necessarily help for the adaptation task. \blue{Finally, on all three applications, \textit{AdaMI} outperforms the two other source-free domain adaptation methods. Specifically, \textit{TTA} yields a smaller improvement than \textit{AdaMI} on the spine and the prostate applications, and fails on the more difficult heart one. \textit{Tent} only yields a small improvement in terms of Dice on all three applications.}

\paragraph{\textcolor{black}{AdaMI versus AdaEnt}}\blue{The Dice scores (DSC) of our proposed \textit{AdaMI} reach $85\%$ of \textit{Oracle}'s performance on spine images, $90\%$ of its performance on prostate images, and $85\%$ on cardiac images. This validates the efficiency of using a class-ratio prior matching with a KL divergence to prevent under-segmentation. Comparing \textit{AdaMI} and \textit{AdaEnt}, we see that on all three applications, \textit{AdaMI} outperforms \textit{AdaEnt} and shows better convergence properties (see Fig.~\ref{fig:diceszfig} (b)). Moreover, in Fig.~\ref{fig:diceszfig} (a), we can observe that \textit{AdaEnt} reaches rapidly its highest validation DSC (first 20 epochs) before slowly decaying. Fig.~\ref{fig:diceszfig} (c) shows that the mean predicted size of structures jumps instantly from 50\% below to 15\% above the mean ground-truth sizes before stagnating}. On the contrary, the performance of \textit{AdaMI} improves steadily and the sizes of predicted structures grow progressively. This suggests that the inversion of the terms in the KL divergence in \textit{AdaMI}, such as in Eq.~\ref{eq:AdaMI}, does help the learning process in domain adaptation, when compared to the original KL divergence in \textit{AdaEnt} (see Section \ref{ssec:invert}). Finally, the ASD values confirm the trend across the different models on cardiac images. Improvement over the lower baseline model (14.6 voxels) is substantial for \textit{AdaEnt} (8.2 voxels), and even greater for \textit{AdaMI} (5.6 voxels), with the greatest improvement occurring for AA and LA structures.

\begin{figure}[t]
  \centering
  \includegraphics[width=0.6\linewidth]{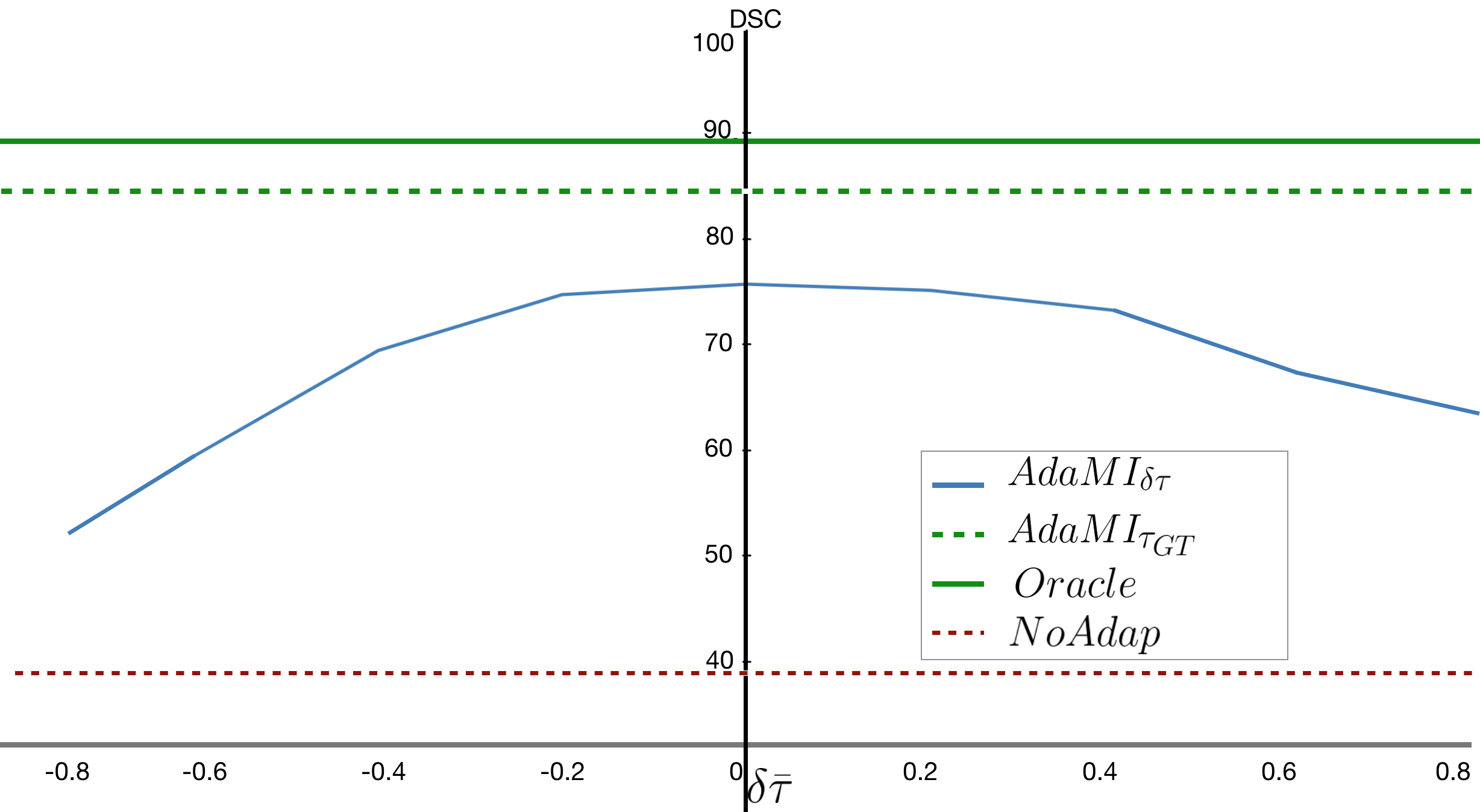}
\caption[]{Robustness performance: DSC (\%) versus enforced relative size error in the class-ratio prior $\delta\bar{\tau}$ for each structure for cardiac segmentation, showing robustness to imprecision in the prior. The DSC performance of the upper bounds \textit{Oracle}, \textit{AdaMI}$_{\tau_{GT}}$ and lower bound \textit{NoAdap} are also indicated.} 
\label{fig:margins}
\end{figure}

\begin{figure}[]
  \centering
  \includegraphics[width=0.6\linewidth]{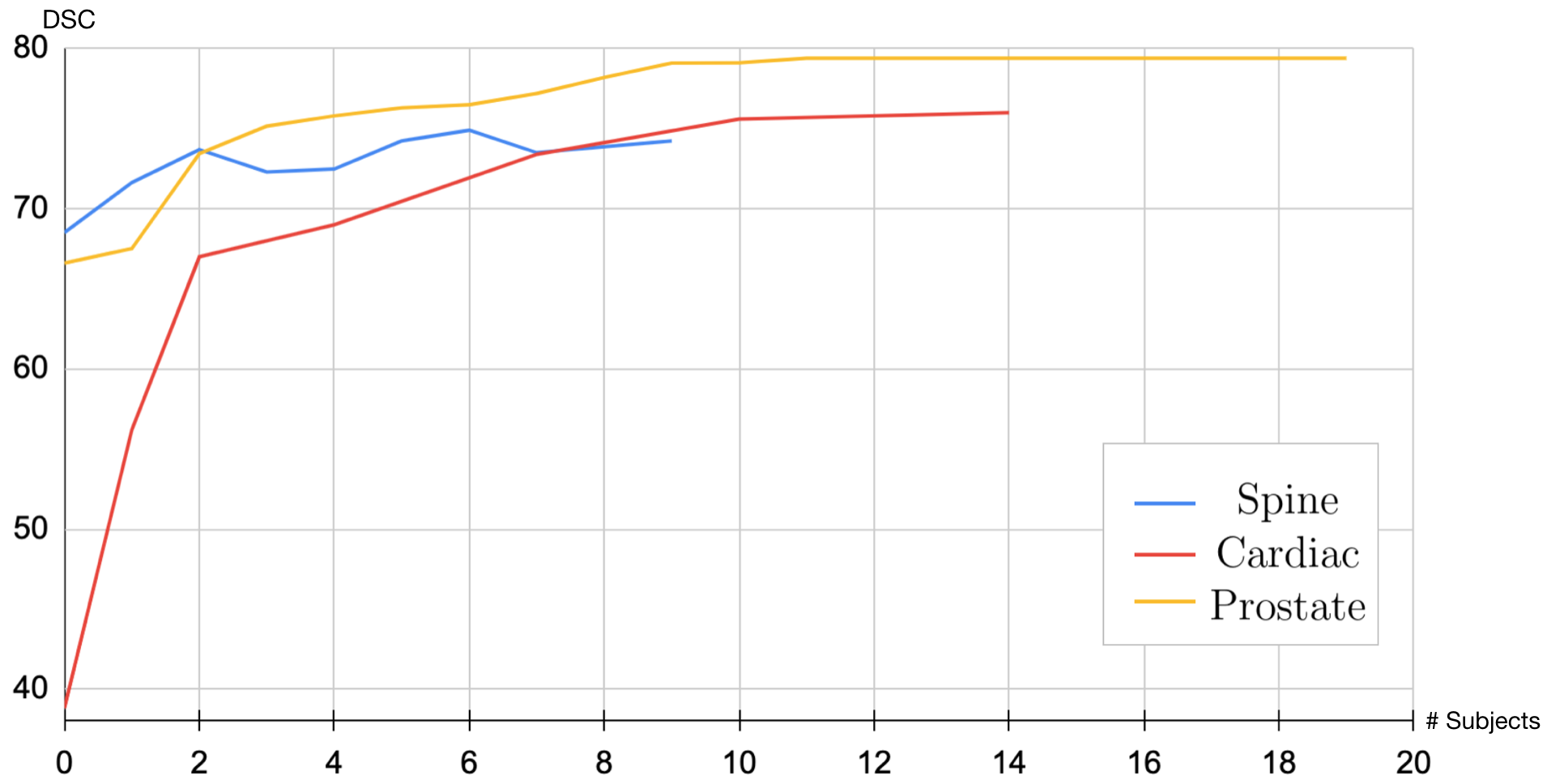}
\caption[]{\blue{Ablation performance: DSC (\%) in target test set versus number of subject in the target training dataset for each application, showing the data efficiency of our method.}} 
\label{fig:dataef}
\end{figure}

\begin{table*}[]
\footnotesize
  \caption{Performance comparison of the proposed formulation with different domain adaptation methods for cardiac segmentation, in terms of DSC (mean) and ASD (mean).}
   \begin{adjustwidth}{0.1cm}{0cm}
   \centering
  \vspace{3mm}
  \setlength\tabcolsep{1.9pt}
  \begin{tabular}{lcccccccc|cccccc}

\toprule

\textbf{Methods} & \textbf{Source}& \textbf{Target} &   \textbf{Backbone}& 
 \multicolumn{5}{c}{\textbf{ DSC (\%)}} & \multicolumn{5}{c}{\textbf{ASD (vox)}} \\
& \textbf{Free}& \textbf{Tags} &   &\textbf{AA} & \textbf{LA} & \textbf{LV} & \textbf{Myo} & \textbf{Mean} & \textbf{AA} & \textbf{LA} & \textbf{LV} & \textbf{Myo} & \textbf{Mean}
\\
\hline
NoAdap (lower bound)&$\checkmark$&$\times$&& 49.8&62.0&21.1&22.1&38.8 & 19.8&13.0&13.3&12.4&14.6  \\
Oracle (upper bound)&$\checkmark$&$\checkmark$&&91.9  & 88.3  & 91.0  & 85.8  & 89.2 & 3.1 & 3.4 & 3.6 & 2.2 & 3.0 \\
\midrule
AdaSource \cite{zhang2019curriculum}&$\times$&$\checkmark$&\multirow{6}{*}{UNet}& 79.0   & 77.9   & 64.4 & 61.3 & 70.7& 6.5 & 7.6 & 7.2 & 9.1 & 7.6 \\
CDA \cite{Bateson2021} & $\times$& $\checkmark$&&77.3 & 72.8 & 73.7 & 61.9 &71.4 & \textbf{4.1} &6.3 &  6.6 &\textbf{6.6} & 5.9 \\
TTA \cite{KARANI2021101907} & $\checkmark$& $\times$&& 59.8 & 26.4 & 32.3& 44.4 & 40.7 & 15.1 & 11.7 & 13.6 &11.3 & 12.9 \\

Tent \cite{wang2021tent} & $\checkmark$& $\times$&& 55.4 & 33.4 &63.0 &41.1 & 48.2 & 18.0 & 8.7 & 8.1 & 10.1 & 11.2 \\
\rowcolor{Gray}Prior AdaEnt \cite{Bateson2020} &$\checkmark$&$\checkmark$&& 75.5&71.2&59.4&56.4&65.6&8.5&7.1&8.4&8.6&8.2\\

\rowcolor{Gray}\textbf{AdaMI (Ours)} &$\checkmark$&$\checkmark$ &&\textbf{83.1}&\textbf{78.2}&\textbf{74.5}&\textbf{66.8}& \textbf{75.7}&5.6&\textbf{4.2}&\textbf{5.7}&6.9&\textbf{5.6}\\
\hline
AdaptSegNet \cite{tsai2018learning} &$\times$& $\times$&  \multirow{6}{*}{DeepLabV2}& 65.4 & 80.6 & 81.4 & 69.3 & 74.2  &8.1 & 5.3 & 4.0 & 3.6 & 5.2 \\
BDL \cite{Li} &$\times$& $\times$&& 67.1 & 80.6 & 82.7 & 62.1 & 73.1& 12.0 & 7.0 & 3.5 & 4.2 & 6.7 \\
CLAN \cite{Luo_2019_CVPR}&$\times$& $\times$&& 63.8 & 79.9 & $\mathbf{84.4}$ & 66.8 & 73.7 &  9.1 & 5.3 & $\mathbf{3.4}$ & $\mathbf{3.5}$ & 5.3 \\
DISE \cite{Chang} &$\times$& $\times$&& 71.8 & 82.2 & 83.7 & 60.8 & 74.6 &6.7 & 4.7 & 3.8 & 7.7 & 5.7 \\
\text{SynSeg-Net \cite{Huo2019}}&$\times$& $\times$&& 71.6 & 69.0 & 51.6 & 40.8 & 58.2 & 11.7 & 7.8 & 7.0 & 9.2 & 8.9 \\
UADA \cite{uada}& $\times$& $\times$&&$\mathbf{84.1}$ &  $\mathbf{88.3}$ &  84.3 &  $\mathbf{71.4}$ & $\mathbf{82.1}$ & $\mathbf{3.9}$ &$\mathbf{3.5}$ &3.8 &$\mathbf{3.7}$ & $\mathbf{3.7}$ \\
\hline
\text{CyCADA \cite{hoffman2017cycada}}&$\times$& $\times$&  \multirow{6}{*}{AdaNet}&72.9 & 77.0 & 62.4 & 45.3 & 64.4 & 9.6 & 8.0 & 9.6 & 10.5 & 9.4 \\
\text{SIFA \cite{SIFA}} &$\times$& $\times$& & \textbf{81.3} & \textbf{79.5} & \textbf{73.8} & \textbf{61.6} & \textbf{74.1} & \textbf{7.9} & \textbf{6.2} & \textbf{5.5} & \textbf{8.5} & \textbf{7.0}\\
\text{PnP-AdaNet \cite{dou2018pnp} }&$\times$& $\times$&  & 74.0 & 68.9 & 61.9 & 50.8 & 63.9 & 12.8 & 6.3 & 17.4 & 14.7 & 12.8 \\
\text{CycleGAN \cite{cyclegan} } &$\times$& $\times$&& 73.8 & 75.7 & 52.3 & 28.7 & 57.6 & 11.5 & 13.6 & 9.2 & 8.8 & 10.8 \\
DANN \cite{dann}& $\times$ & $\times$&&39.0& 45.1& 28.3&25.7& 34.5 &16.2&9.2&12.1&10.1&11.9\\
ADDA \cite{ADDA} & $\times$ & $\times$ && 47.6&60.9& 11.2&29.2 & 37.2&13.8&10.2& NA& 13.4&NA \\
\hline

Overall ranking of AdaMI (\#/16) &&&& 2&7&6&3&\textbf{2}&3&	2&7&6&\textbf{4}\\
\bottomrule\\[-8pt]
  \end{tabular}
  \label{table:resultswhs2}
   \end{adjustwidth}
\end{table*}

\subsection{Ablation study on class-ratio precision}  
\label{ssec:ablaratio}
We also investigate the impact of imprecision in the target domain class-ratio prior on the quality of SFDA models. To this end, we validate a range of values in the estimations of class-ratios, as explained in Sec.~\ref{sssec:robustness}. The results are reported for cardiac images in Fig.~\ref{fig:margins}. First, in the ideal situation where the precise class-ratios are known, ${AdaMI_{\tau_{GT}}}$ reaches 84.5\% DSC, representing 95\% of the upper baseline, the \textit{Oracle}. Then, we can see that our proposed method \textit{AdaMI} is robust to large ranges of imprecision in class-ratio estimates. Indeed, a difference of $\pm$ 20\% (resp. $\pm$ 40\%) with our prior estimation in Sec.~\ref{ssec:sizeprior} only degrades the DSC by up to 1\% (resp. 6\%).
\textcolor{black}{Moreover, we see that an overestimation of the structure sizes leads to a better overall DSC than an underestimation, highlighting the well-known bias of Dice towards over-segmentation.}



Finally, we emphasize that the class-ratio estimation used for a structure $k$ is identical for all target images containing $k$. However, the true target class-ratios have high variance amongst slices. Thus the prior used in \textit{AdaMI} is very coarse, which further confirms the robustness of our framework to class-ratio prior imprecision.


\subsection{\blue{Ablation study on the size of the target training dataset}}  \label{ssec:ablationtts}

\blue{We also investigate how much weakly-labeled target training data is necessary for our SFDA model to achieve adaptation. To this end, we experiment  with a varying number of subjects in the target training dataset. The results are reported in Fig.~\ref{fig:dataef}. We can see that our proposed method \textit{AdaMI} is robust to large diminution of target dataset size. Indeed, with only 2 subjects,  \textit{AdaMI} is on par with most state-of-the art methods, reaching 67\% DSC for the cardiac application, 74\% DSC for the spine, and 73\% for the prostate.}

\begin{table}[t]
\centering
\footnotesize
\caption{\blue{Performance of the proposed formulation obtained when removing the weak image-level annotations.}}

\setlength\tabcolsep{5pt} 
\begin{tabular}{lccccccc}
\toprule
\textbf{Method} & \textbf{Target Tags} & \textbf{Dataset} &\textbf{DSC}  & \textbf{ASD}  \\
\midrule
\multirow{3}{*}{\begin{tabular}[c]{@{}l@{}}$AdaMI_{unsupervised}$\end{tabular}} &\multirow{3}{*}{\begin{tabular}[c]{@{}l@{}}$\times$\end{tabular}} & IVDM3Seg
   & 73.7  & 1.33 \\
&&NCI-ISBI13
   & 71.8 & 7.49  \\
&&  MMWHS
   & 58.0& 12.2  \\
 \bottomrule\\[6pt]
\end{tabular}
\label{table:resultsnotag}
\end{table}

\subsection{\blue{Ablation study on weak the annotations in the target training dataset}}  \label{ssec:ablationilt}

\blue{Finally, we investigate the more general scenario where images are fully unsupervised in the target domain. Particularly, we removed the target image tags for the adaptation phase as explained in Section \ref{ssec:sizeprior}. Results from this study are reported in Table \ref{table:resultsnotag}. As expected, having image-level tag information helps to all the models, which can be observed from the performance degradation compared to results in Table \ref{table:resultsivd} and \ref{table:resultswhs2}. Indeed, the class-ratio estimation degrades without the image tag, and as a result, models using a class-ratio prior to guide adaptation also see their performance decrease.} \textcolor{black}{However, for the spine and the prostate application, the quantitative performance (73.7\% DSC and 71.8\% DSC respectively) remains well above the baseline, on par with most state-of-the art domain adaptation models. The removing of image-level Tags is more difficult for the heart application, as it is multi-class and has a big domain shift. However, results (58.0\% DSC) stayed well above both the baseline and the two other SFDA methods, \textit{Tent} and \textit{TTA}.}

\subsection{Qualitative results}
\begin{figure*}[t]
\centering
    \includegraphics[width=0.5\linewidth]{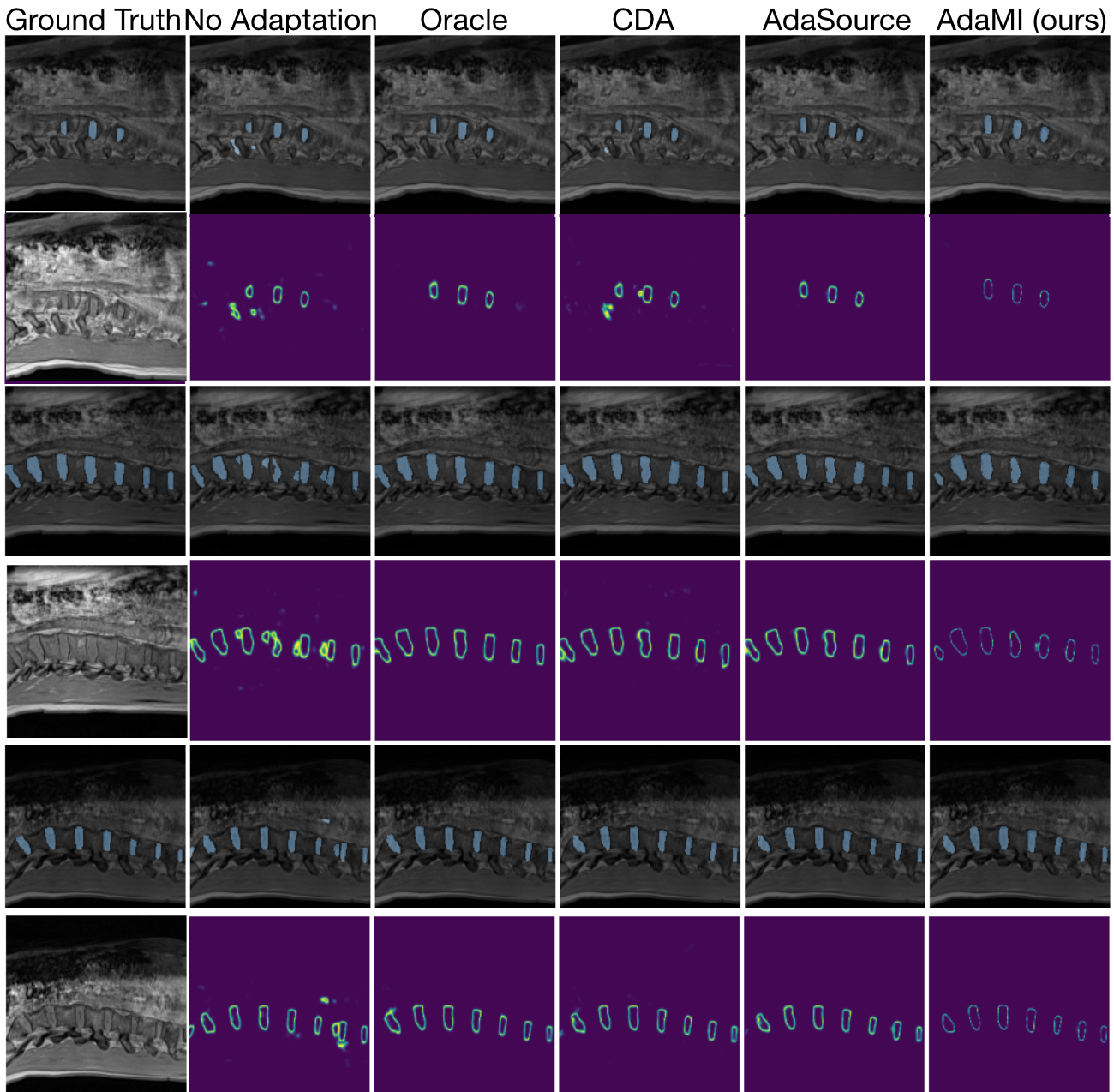}
      \caption[]{Qualitative performance on spine MRI images: examples of the segmentations achieved by our formulation (\textit{AdaMI}), benchmark models in \cite{Bateson2021}, \cite{zhang2019curriculum} and lower (\textit{NoAdap}) and upper baselines (\textit{Oracle}). First column shows an input slice and the corresponding semantic segmentation ground-truth. The other columns show segmentation results (top) along with prediction entropy maps produced by the different models (bottom). }
         \label{fig:segmivd}
\end{figure*}

\begin{figure*}[t]
\centering
    \includegraphics[width=0.5\linewidth]{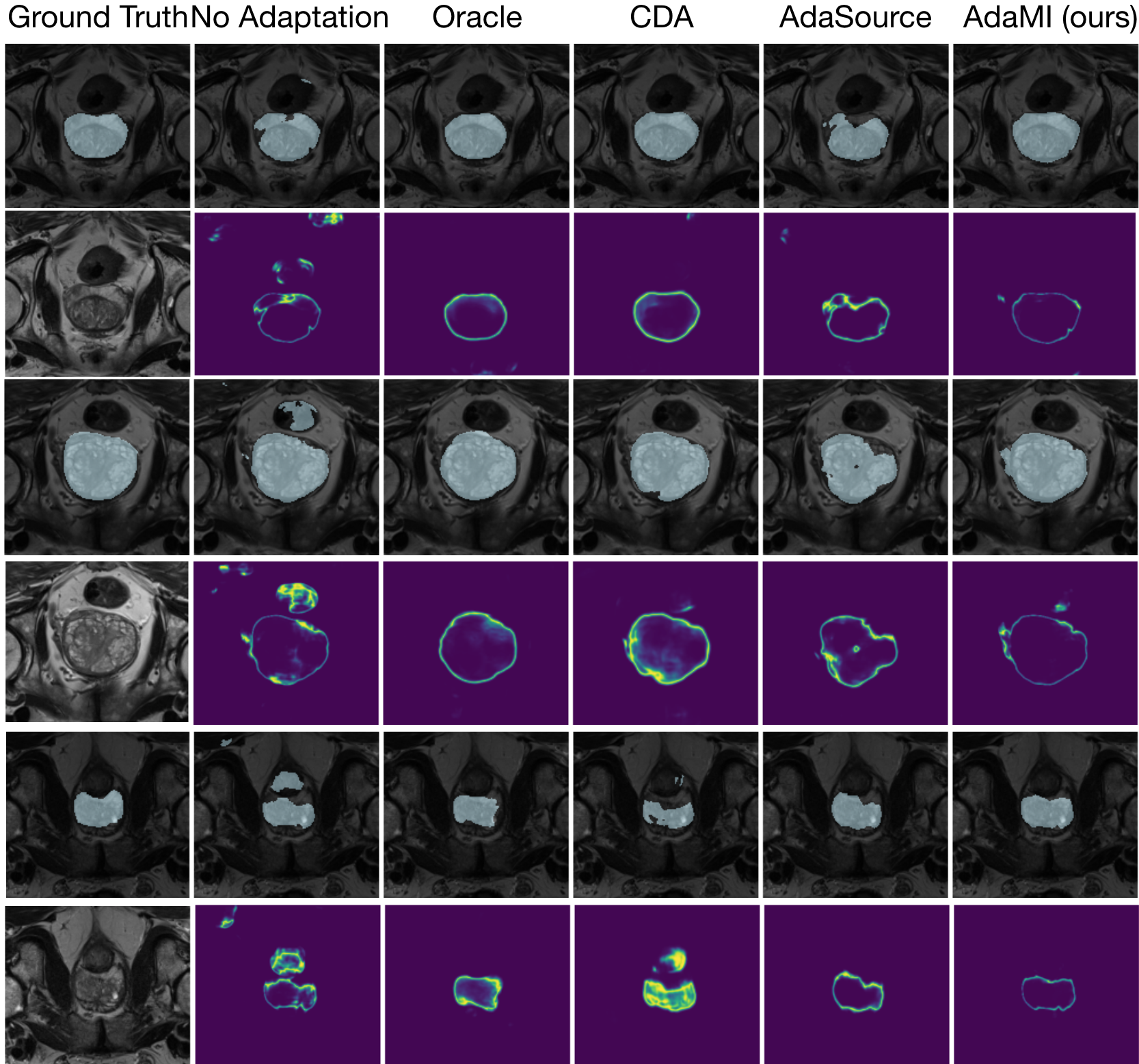}
      \caption[]{Qualitative performance on prostate MRI images: examples of the segmentations achieved by our formulation (\textit{AdaMI}), benchmark models in \cite{Bateson2021}, \cite{zhang2019curriculum} and lower (\textit{NoAdap}) and upper baselines (\textit{Oracle}).}
         \label{fig:segmpro}
\end{figure*}

\begin{figure*}[t]
\centering
\includegraphics[width=0.5\linewidth]{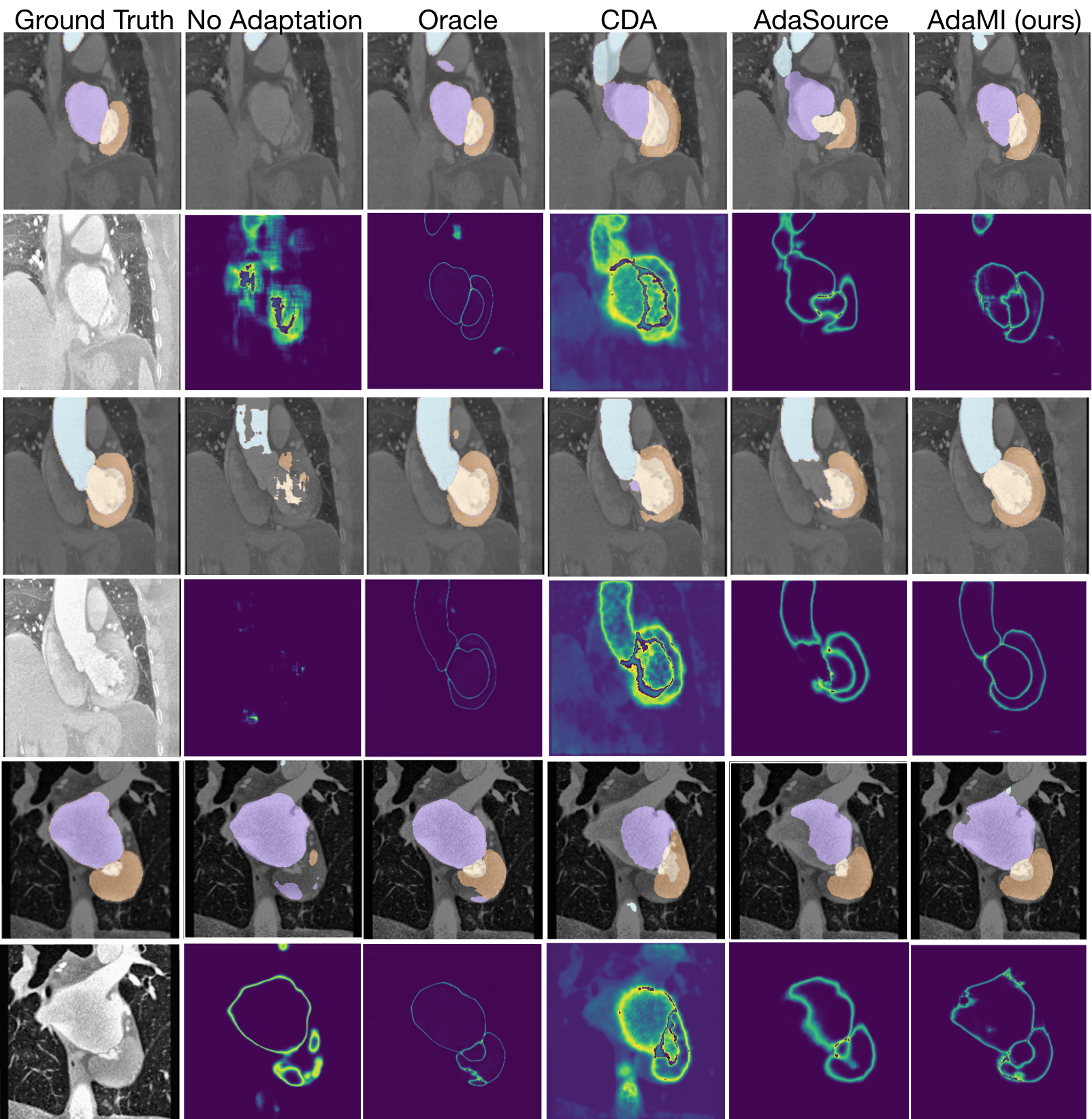}
\vspace{-1mm}
\caption{Qualitative performance on cardiac CT images: examples of the segmentations achieved by our formulation (\textit{AdaMI}), benchmark models in \cite{Bateson2021}, \cite{zhang2019curriculum} and lower (\textit{NoAdap}) and upper baselines (\textit{Oracle}). The cardiac structures of MYO, LA, LV and AA are depicted in brown, purple, yellow and blue, respectively.}
\label{fig:segmwhs}
\end{figure*}

Qualitative segmentations and the corresponding entropy maps are shown for spine images in Fig.~\ref{fig:segmivd}, for prostate images in Fig.~\ref{fig:segmpro}, and for cardiac ones in Fig.~\ref{fig:segmwhs}. Without adaptation, the predictions of the network are either uncertain, as revealed by the high activation in the entropy maps of predictions (see top two lines in Fig.~\ref{fig:segmwhs}); or severely biased towards the dominant class, i.e. the background. This bias produces under-segmented or completely undetected structures (see the top four rows in Fig.~\ref{fig:segmwhs}). In all cases, the output segmentation masks are noisy, with very irregular edges.  
Benchmark adaptation models \textit{CDA} and \textit{AdaSource} are able to recover the structures in most examples. However, they display high uncertainty in the predictions, especially \textit{CDA}. Interestingly, for some difficult slices, the segmentation results produced by our proposed SFDA model matches better with the ground-truth. For spine and prostate images, such examples are displayed in bottom two rows in Fig.~\ref{fig:segmpro}. For cardiac images, the whole AA structure is better recovered (see middle two rows in Fig.~\ref{fig:segmwhs}), and the shapes and the boundary between the MYO and the LV structure are improved. Notably, in all applications, the entropy maps produced by \textit{AdaMI} only show high activations along the borders of the predicted structures. These visual results further confirm the remarkable ability of \textit{AdaMI} to produce accurate predictions with high confidence over existing approaches. 

\section{Discussion}
\label{sec:discussion}

We have introduced a source-free domain adaptation (SFDA) method to guide a segmentation network, trained on a source domain, to perform on a different target domain, without any access to the source-domain data in the adaptation phase. We have demonstrated the robustness of our SFDA approach on cross-modality spine MRI, cross-site prostate MRI, and MRI-to-CT cardiac adaptation.

\textbf{Source-Free Domain Adaptation:}  Surprisingly, even though our model does not access the source data in the adaptation phase, it yields comparable or better performance than many state-of-the art adaptation approaches that do rely on the source data. \blue{It also outperforms two very recent source-free domain adaptation approaches, \cite{wang2021tent,KARANI2021101907}. These works have stressed on the need for limited flexibility at test time, by freezing most parameters in the network, and adapting only the normalization and affine ones. Yet, in our three applications, we have found our proposed method, where the entire network is adapted, to be more efficient.} Furthermore, our principled solution to source-free domain adaptation minimizes the uncertainty of the target domain predictions while preventing trivial solutions of single-class outputs via a KL regularizer that encourages target class-ratio (i.e region proportions). Using entropy minimization in combination with this regularizer, our formulation reaches 85\%, 90\% and 85\% of full supervision in spine, prostate, and cardiac images respectively. Our qualitative results demonstrate the ability of SFDA to produce accurate predictions with high confidence. 

\textbf{Robustness:} Our experiments have further confirmed the robustness of \textit{AdaMI} to substantial prior imprecision, and that having a coarse knowledge of the target region proportions can be enough to guide adaptation. In our implementation, a class-ratio prior is derived from readily available anatomical reference values. This anatomical knowledge is combined with image-level tags to produce a very coarse yet effective estimation of target class-ratios. This finding has great potential value in the medical domain, as prior anatomical knowledge is commonly available, due to conventions in patient position and anatomical similarity \cite{jurdi2020highlevel}. We have, therefore, proposed an effective method to integrate such domain-invariant knowledge, with straightforward extensions in many medical applications.

We have also shown that, in the ideal setting when a very precise prior is known, the performance of \textit{AdaMI} is close to full supervision. This suggests that \textit{AdaMI} is able to approach the \textit{''optimal'' segmentation given the amount of prior imprecision}. This finding is in line with the recent work of \cite{KervadecMIDL2021}, which shows that using a few global shape descriptors as supervision enables performances close to a full pixel-wise supervision. In fact, the class-ratio used in \textit{AdaMI} is based on zero-order shape moments. 

We have also demonstrated the superiority of \textit{AdaMI} when compared to our previous \textit{AdaEnt}, which regularizes the class-ratio priors with a steeper loss \cite{Bateson2020}. 
Indeed, \textit{AdaMI} is able to prevent the under-segmentation regime observed without adaptation, while avoiding the fast convergence to local minima observed with \textit{AdaEnt}. Although convergence and stability are well-known challenges for unsupervised and weakly supervised deep domain adaptation methods, \textit{AdaMI} shows remarkable training stability. 
On cross-modality spine MRI and cross-site prostate MRI, our method has shown performances on par with several adaptation models that necessitate both the source and target data, such as \cite{Bateson2021,zhang2019curriculum}. Surprisingly, for the adaptation of MRI-to-CT cardiac images, our model outperforms several recent state-of-the-art adaptation models, such as \cite{Bateson2021,zhang2019curriculum,tsai2018learning,SIFA,hoffman2017cycada,cyclegan,ADDA,dann}. This is confirmed qualitatively by our experiments, where the structures of interests are well predicted in all the three applications. In some cases, the segmentation masks are even improved when compared to benchmark adaptation models, despite the lack of source data. These results, therefore, suggest that having access to the source data may not be necessary for domain adaptation.

\blue{\textbf{Extension to 3D:} In our experiments on all three applications, the images are 3D volumes. As we have used a standard 2D segmentation network (2D-UNet \cite{UNet}), we input slices from these 3D volumes for training and inference. However, our method can be extended to be fully-3D; to this end, 3D class-ratio priors should be obtained, to adapt a 3D segmentation network (such as 3D-UNet \cite{3DUNET}).}



\textbf{Limitations:}  \textcolor{black}{A limitation of our work is the need for an image-level annotation, compared to fully unsupervised domain adaptation methods. However, the majority of these domain adaptation methods use both the source and target data, and are much more complex. Very recent test-time domain adaptation methods such as \cite{He2020,KARANI2021101907} also comply with the source-free domain adaptation scenario, but at the cost of an auxiliary branch or additional training tasks in source training phase.} Instead, our method tackles the adaptation problem with no alteration in the source training phase, by optimizing a single network, and uses only the target images in the adaptation phase. Importantly, this drastically reduces the computational burden, while easing optimization difficulty, when compared to state-of-the-art domain adaptation models, notably adversarial methods. \textcolor{black}{Indeed, these methods rely on a two-step training of two networks, a discriminator and a segmenter, and a dependency on data from both the source and target domains. Another limitation, robustness to a large shift of class-ratio distributions compared to anatomical reference values (e.g. population-wise differences), could be challenging for out method. However, we show in Table that our model large ambiguities ($\pm 60\%$) on these class-ratios distributions only degrade the performance by up to 15\%.} 


\section{Conclusion}

\textcolor{black}{Our proposed Source-Free Domain Adaptation (SFDA) tackles a source-free domain adaptation for semantic segmentation, which removes the need for a concurrent access to the source and target data during adaptation.} Our approach substitutes the standard supervised loss in the source domain by a direct minimization of the entropy of predictions in the target domain. To prevent trivial solutions, we regularize the entropy loss with a class-ratio prior, which is derived from approximate anatomical knowledge.
Unlike recent domain-adaptation techniques, our method tackles domain adaptation without resorting to source data during the adaptation phase, a setting of great value in practice. Interestingly, our formulation achieves a better performance than several state-of-the-art methods which still need access to both source and target data. Our source-free approach has been validated with cross-modality intervertebral discs segmentation, cross-site prostate segmentation and MRI to CT cardiac substructure segmentation.
This shows the effectiveness of our prior-aware entropy minimization and that adaptation might not need access to the source data, even when the domain shift is large, as suggested by our experiment on MR to CT cardiac images. 
Future work will address the integration of other anatomical priors. Our proposed adaptation framework is straightforward to use, drastically reduces the computational burden of the domain adaptation, the optimization complexity, and can be used with any segmentation network architecture.

\section*{Acknowledgments}
This work is supported by the Natural Sciences and Engineering Research Council of Canada (NSERC), Discovery Grant program, by the The Fonds de recherche du Québec - Nature et technologies (FRQNT) grant, the Canada Research Chair on Shape Analysis in Medical Imaging, the ETS Research Chair on Artificial Intelligence in Medical Imaging, and NVIDIA with a donation of a GPU. The authors would like to thank the MICCAI 2018 IVDM3Seg, NCI-ISBI 2013, and MICCAI 2017 MMWHS organizers for providing the data.
\bibliographystyle{splncs04}
\bibliography{biblio}


\newpage

\appendix
\section{Estimation of the class-ratio priors from anatomical knowledge}\label{appendix}

We detail below the estimation of the class-ratio priors for each application. Note that for a structure k, after obtaining the estimated size in $mm^{2}$, the class-ratio (i.e. region proportion) $\tau_k$ is calculated as : $\bar{\tau}_k=\frac{size_k}{R_1*R_2*\Omega}$, where $R_1$ and $R_2$ are the resolution values in the corresponding plane ($R_1$ = $R_2$ when isotropic)  and $\Omega$ is the cardinal size of the image. Table \ref{tab:estimatedsizes} summarizes the estimations obtained for each structure.

\paragraph{IVDM3Seg}
Monitoring Lumbar intervertebral disc dimensions is useful treat lumbar spine diseases and for surgical reconstructions. Reference average Lumbar disc height $H$ was available in \cite{IVD2}, and the antero-posterior Lumbar disc diameter $D$ was available in \cite{IVD3}. The area for one IVD is obtained as $size_{IVD}=\frac{\pi\times H \times D}{4}$, and the final area is $size_{IVDS}=7\times size_{IVD}$.

\paragraph{NCI-ISBI13}
Prostate volume and dimensions are widely monitored. Reference volume $V$ and height $H$ were taken from \cite{Prostate}, which measured them through planimetry. We calculated the transverse surface dimension as: $size_{Prostate}=\frac{3V}{2H}$.

\paragraph{MMWHS} \footnote{\label{note1}As we used the preprocessed data from \cite{dou2018pnp}, which had performed cropping, zooming and resampling of the slices, we estimated the resolution of these preprocessed slices in the coronal plane as $0.45\times0.93\mathrm{~mm} / \mathrm{px}$}
\textbf{LA} Reference LA area dimensions are readily available as LA area is a useful biomarker in clinical assessment of heart diseases. We used the measurement in \cite{LeftAtrium} Table 1, taken at maximum volume (end-systole) in the 4-chamber view\footnote{\label{note2}Note that these planes are slightly different from the coronal imaging plane of the cardiac slices used in our framework, leading to imprecisions in our estimations.}; \textbf{LV} In \cite{LV} Table 3, an estimation of LV area is computed by averaging measurements across 12 long-axis angles\textsuperscript{\ref{note1}}. We took the measurement at maximum volume, i.e. end-diastole, to estimate $size_{LV}$; \textbf{AA} We used aortic diameters at proximal (p) and distal (d) levels as given in \cite{AA}, as well as the average AA length (l) provided by the MMWHS organisers \footnote{http://www.sdspeople.fudan.edu.cn/zhuangxiahai/0/mmwhs/data.html} to calculate an estimation of the average AA area in a coronal slice as: $size_{AA}=\frac{p+d}{2}*l+\pi *(p/4)^{2}$. \textbf{Myo} The Myo is the structure with the most complicated geometry, thus obtaining an accurate estimation of $size_{Myo}$ is difficult. However, in \cite{Myo}, left ventricular myocardial and cavity volumes are available at end-diastole (LVEDV and MVd respectively) and end-systole (LVESV and MVs). We calculate these two ratios : $r_{diastole}=\frac{LVEDV}{MVd}$ ; $r_{systole}=\frac{LVESV}{MVs}$ and estimate the average Myo area in a coronal slice as : 
$size_{AA}=\frac{r_{diastole}+r_{systole}}{2}*size_{LV}$.

\begin{table}[]
\centering
\scriptsize
\begin{tabular}{lc|c|cccc}
&\multicolumn{1}{c}{\textbf{IVDM3Seg}}&\multicolumn{1}{c}{\textbf{NCI-ISBI13}} &\multicolumn{4}{c}{\textbf{ MMWHS}} \\
  &IVD&Prostate& Myo & LA & LV & AA \\
\midrule
$size_k$ (mm2) & 2784 & 2485& 1871 & 2110  & 1895 & 1565 \\
$size_k$ (pix) & 1782 &6095& 4428 & 4996 & 4487 & 3706 \\
$\bar{\tau}_k$ (\%) & 2.72 &4.68& 6.76 & 7.62 & 6.85 & 5.65 \\
\bottomrule
\end{tabular}
    \caption[]{Estimated sizes and class-ratios of structures in the target datasets}
    \label{tab:estimatedsizes}
\end{table}

\section{Link between the loss in \textit{AdaMI} and mutual information maximization}\label{miproof}

Given the following expression of the mutual information between two random variables \textit{X} and \textit{Y}:

\begin{equation*}
\begin{aligned}
\mathcal{I}(X ; Y) &=\mathbb{E}_{Y}\left[\log \mathbb{E}_{X}[p(Y \mid X)]\right]-\mathbb{E}_{X, Y}[\log p(Y \mid X)]
\end{aligned}
\end{equation*}

The mutual information between an input image $I_t$ and its softmax predictions $P_t$ can be written as:

\begin{equation*}
    \mathcal{I}(I_t ;P_t) = \mathbb{E}_{P_t}\left[\log \mathbb{E}_{I_t}[p(P_t \mid I_t)]\right] -\mathbb{E}_{I_t,P_t}[\log p(P_t \mid I_t)]
\end{equation*}

And recall that $\mathbb{E}_{I_t}[p(P_t \mid I_t)] = \frac{1}{\left | \Omega_t \right |} \sum_{i \in \Omega_t} {\mathbf p}_t (i, \theta) = \hat{\tau}(t,.,\theta)$.  Decomposing for each term, \blue{and assuming pixel-wise independence of $P_t$}, we obtain:

\begin{equation*}
\begin{aligned}
\mathbb{E}_{P_t}\left[\log \mathbb{E}_{I_t}[p(P_t \mid I_t)]\right] &=\mathbb{E}_{P_t}\left[\log \hat{\tau}(t,.,\theta)\right] \\
&=\frac{1}{\left|\Omega_{t}\right|}\sum_{i \in \Omega_t}\sum_{k=1}^{K}p^{k}_t(i,\theta)\log \hat{\tau}(t,k,\theta) \\
&=\sum_{k=1}^{K}\hat{\tau}(t,k,\theta)\log \hat{\tau}(t,k,\theta)=-H\{\hat{\tau}(t,.,\theta)\}
\end{aligned}
\end{equation*}

and : 

\begin{equation*}
\begin{aligned}
-\mathbb{E}_{I_t,P_t}[\log p(P_t \mid I_t)] 
&=-\frac{1}{\left|\Omega_{t}\right|}\sum_{i \in \Omega_t}\sum_{k=1}^{K}p^{k}_t(i,\theta)\log p^{k}_t(i,\theta) \\
&=\frac{1}{\left|\Omega_{t}\right|}\sum_{i \in \Omega_t} \ell_{ent}({\mathbf p}_t (i, \theta))\}
\end{aligned}
\end{equation*}

The following identity follows:

\begin{equation*}
\mathcal{I}(I_t ;P_t)=\underbrace{-H\{\hat{\tau}(t,.,\theta)\}}_{\mathbb{E}_{P_t}\left[\log \mathbb{E}_{I_t}[p(P_t \mid I_t)]\right]} + \underbrace{\frac{1}{\left|\Omega_{t}\right|}\sum_{i \in \Omega_t} \ell_{ent}({\mathbf p}_t (i, \theta))}_{-\mathbb{E}_{I_t,P_t}[\log p(P_t \mid I_t)]}
\end{equation*}

Finally the empirical estimation of the mutual information between a set of input images $I_t$ and their latent label predictions $P_t$, $t=1 \dots T$ is given by:

\begin{equation*}
\mathcal{I_\theta} = \frac{1}{T}\sum_{t=1}^{T}\mathcal{I}(I_t ;P_t) = \frac{1}{T}\sum_{t=1}^{T}-H\{\hat{\tau}(t,.,\theta)\} + \frac{1}{\left|\Omega_{t}\right|}\sum_{i \in \Omega_t} \ell_{ent}({\mathbf p}_t (i, \theta))
\end{equation*}


\end{document}